\documentclass[conference]{IEEEtran}
\IEEEoverridecommandlockouts
\usepackage{cite}
\usepackage{amsmath,amssymb,amsfonts}
\usepackage{algorithmic,algorithm}
\usepackage{graphicx}
\usepackage{textcomp}
\usepackage{xcolor}
\usepackage{stackengine}
\usepackage{subfig}
\usepackage{hyperref}
\usepackage{balance}

\usepackage{caption}
\def\BibTeX{{\rm B\kern-.05em{\sc i\kern-.025em b}\kern-.08em
    T\kern-.1667em\lower.7ex\hbox{E}\kern-.125emX}}
\begin{document}

\title{SeriesGAN: Time Series Generation via Adversarial and Autoregressive Learning}

\author{
    \IEEEauthorblockN{MohammadReza EskandariNasab, Shah Muhammad Hamdi, Soukaina Filali Boubrahimi}
    \IEEEauthorblockA{
        \textit{Department of Computer Science, Utah State University, Logan, UT 84322, USA} \\
        Emails: \{reza.eskandarinasab, s.hamdi, soukaina.boubrahimi\}@usu.edu \\
        ORCID: {0009-0004-0697-3716}, {0000-0002-9303-7835}, {0000-0001-5693-6383}
    }
}

\maketitle

\begin{abstract}
Current Generative Adversarial Network (GAN)-based approaches for time series generation face challenges such as suboptimal convergence, information loss in embedding spaces, and instability. To overcome these challenges, we introduce an advanced framework that integrates the advantages of an autoencoder-generated embedding space with the adversarial training dynamics of GANs. This method employs two discriminators: one to specifically guide the generator and another to refine both the autoencoder's and generator's output. Additionally, our framework incorporates a novel autoencoder-based loss function and supervision from a teacher-forcing supervisor network, which captures the stepwise conditional distributions of the data. The generator operates within the latent space, while the two discriminators work on latent and feature spaces separately, providing crucial feedback to both the generator and the autoencoder. By leveraging this dual-discriminator approach, we minimize information loss in the embedding space. Through joint training, our framework excels at generating high-fidelity time series data, consistently outperforming existing state-of-the-art benchmarks both qualitatively and quantitatively across a range of real and synthetic multivariate time series datasets.
\end{abstract}

\begin{IEEEkeywords}
Time Series Generation, Generative Adversarial Networks, Autoregressive Models, Autoencoders, Data Augmentation
\end{IEEEkeywords}

\section{Introduction}
\label{sec:intro}

Generating realistic synthetic data can help balance datasets and mitigate data shortages \cite{eskandariNasab2024solarflare}, thus enhancing scientific research and boosting the effectiveness of various machine learning applications. However, generating time series data presents unique challenges due to its temporal characteristics \cite{eskandarinasab2024grucnn}. Models need to capture not only the distribution of features at each time point but also the complex interactions between these points over time. For example, in multivariate sequential data \(\mathbf{x}_{1:T} = (\mathbf{x}_1, \ldots, \mathbf{x}_T)\), a good model should accurately determine the conditional distribution \(p(\mathbf{x}_t \mid \mathbf{x}_{1:t-1})\) to reflect temporal transitions. This capability is crucial across numerous fields, especially when working with imbalanced time series datasets \cite{angryk2020multivariate}. Areas such as healthcare \cite{eskandarinasab2024grucnn} and solar physics \cite{hamdiflare, vural2024contrastive} often encounter data limitations due to factors such as privacy issues, the complexity and noise in data, or the rarity of events, which make model training and evaluation challenging. By developing approaches that utilize generative adversarial networks (GANs) \cite{goodfellow2014generative} to generate realistic synthetic data, scientific advancement can be supported, and machine learning performance can be enhanced by balancing datasets and addressing data scarcity \cite{eskandariNasab2024solarflare}.

A significant amount of research has focused on improving the temporal dynamics of autoregressive models for sequence forecasting, aiming to minimize the impact of sampling errors by making various adjustments during training to better model conditional distributions \cite{b4, b5}. Autoregressive models break down the sequence distribution into a series of conditionals \(\prod_{t} p(\mathbf{x}_t \mid \mathbf{x}_{1:t-1})\), making them effective for forecasting due to their deterministic properties. However, these models are not genuinely generative, as they do not require external input to generate new sequences. On the other hand, research involving GANs for sequential data typically employs recurrent networks as generators and discriminators, aiming directly at an adversarial objective \cite{mogren2016crnngan, esteban2017realvalued, ramponi2019tcgan}. While this method is straightforward, the adversarial objective targets modeling the joint distribution \(p(\mathbf{x}_{1:T})\) without accounting for the autoregressive nature, which might be insufficient since merely aggregating standard GAN losses over vectors may not adequately capture the stepwise dependencies present in the training data.

In this paper, we propose a new framework that substantially improves the stability, quality of generated data, and generalizability. Our approach, named \emph{SeriesGAN}, seamlessly integrates two research domains, GANs and autoregressive models, into a powerful and accurate generative model uniquely tailored to preserve temporal dynamics. SeriesGAN offers a holistic solution for generating realistic time-series data, with broad applicability across multiple fields. The primary contributions of our work include:

\begin{enumerate}
    \item Utilizing two discriminators (dual-discriminator training) that operate separately on feature and latent spaces, providing dual feedback for the generator. The feedback from the feature space also assists the autoencoder in enhancing its reconstruction capability and accuracy.
    \item Developing a novel autoencoder-based loss function for the generator network, which enhances the quality of the generated data and facilitates optimal convergence. Additionally, a new loss function is designed for the autoencoder network.
    \item Employing a teacher-forcing-based supervisor network with a novel loss function, which significantly helps the generator network to better learn the temporal dynamics of time-series data.
    \item Implementing an early stopping algorithm and applying Least Squares GANs (LSGANs) \cite{mao2017squares} to stabilize the framework and ensure optimal results at the end of each training session.
\end{enumerate}

We demonstrate the advantages of SeriesGAN by conducting a series of experiments on a variety of real-world and synthetic multivariate and univariate time series datasets. Our findings indicate that SeriesGAN consistently outperforms existing benchmarks, including TimeGAN \cite{NEURIPS2019_c9efe5f2}, in generating realistic time-series data.

\section{Related Work}
\label{sec:relatedwork}

Autoregressive recurrent networks trained using maximum likelihood methods tend to experience substantial prediction errors during multi-step sampling \cite{Williams1989}. This problem arises due to the discrepancy between \textit{closed-loop training} (conditioned on real data) and \textit{open-loop inference} (based on previous predictions). Drawing inspiration from adversarial domain adaptation \cite{Pforcing}, Professor Forcing employs an additional discriminator to distinguish between autonomous and teacher-driven hidden states \cite{Tforcing}, thereby aligning training and sampling dynamics. Teacher forcing reduces errors during training by using ground truth data for conditioning at each step, while Professor Forcing bridges the gap between training and inference by aligning the hidden state dynamics between both processes. These methods aim to reduce exposure bias, which occurs when a model’s predictions degrade over time during inference. However, despite these methods aiming to model stepwise transitions, they are deterministic and do not explicitly involve sampling from a learned distribution, which is essential for our goal of synthetic data generation.

The seminal work on GANs \cite{goodfellow2014generative} presented a groundbreaking framework for generating synthetic data. This model comprises two neural networks (a generator and a discriminator) trained concurrently within a zero-sum game structure. The generator learns to produce data by attempting to fool the discriminator, while the discriminator simultaneously learns to distinguish between real and generated data. Both networks improve iteratively through this adversarial process, with the generator striving to minimize the discriminator’s accuracy. In many GAN implementations, CNNs \cite{cnn} are employed to enhance the generator and discriminator’s ability to capture spatial patterns. While GANs can generate data by sampling from a learned distribution, they face challenges in capturing the sequential dependencies characteristic of time series data. The adversarial feedback from the discriminator alone does not provide enough information for the generator to adequately learn the intricate patterns within sequences.

Several studies have employed the GAN framework for time series analysis. The earliest of these, C-RNN-GAN \cite{mogren2016crnngan}, applied the GAN architecture directly to sequential data, using LSTM networks as both the generator and discriminator. This model generates data recurrently, starting with a noise vector and the data from the previous time step. RCGAN \cite{esteban2017realvalued} improved upon this by eliminating the dependence on previous outputs and incorporating additional inputs for conditioning \cite{mirza2014conditional}. However, unlike TimeGAN, these models depend solely on binary adversarial feedback for learning, which may not sufficiently capture the temporal dynamics of the training data. TimeGAN \cite{NEURIPS2019_c9efe5f2} offers an advanced method for generating realistic time series data by merging the flexibility of unsupervised learning with the accuracy of supervised training. It leverages an autoencoder, enabling the GAN to both generate and discriminate within the latent space. This approach helps the GAN mitigate non-convergence issues \cite{nonconverge} by training in a lower-dimensional representation. TimeGAN is designed to accurately replicate the temporal dynamics inherent in training data, making it particularly useful for addressing imbalanced time series classification problems such as solar flare prediction, where X-class flares are rare occurrences. Employing data augmentation techniques such as TimeGAN can help boost the predictive performance of solar flare prediction models \cite{eskandarinasab2024enhancing}. However, despite its advantages, it struggles with stability during training and often produces inconsistent data quality. As a result, it frequently fails to deliver optimal results after each training cycle.

\begin{figure}
\centering
\includegraphics[width=0.38\textwidth]{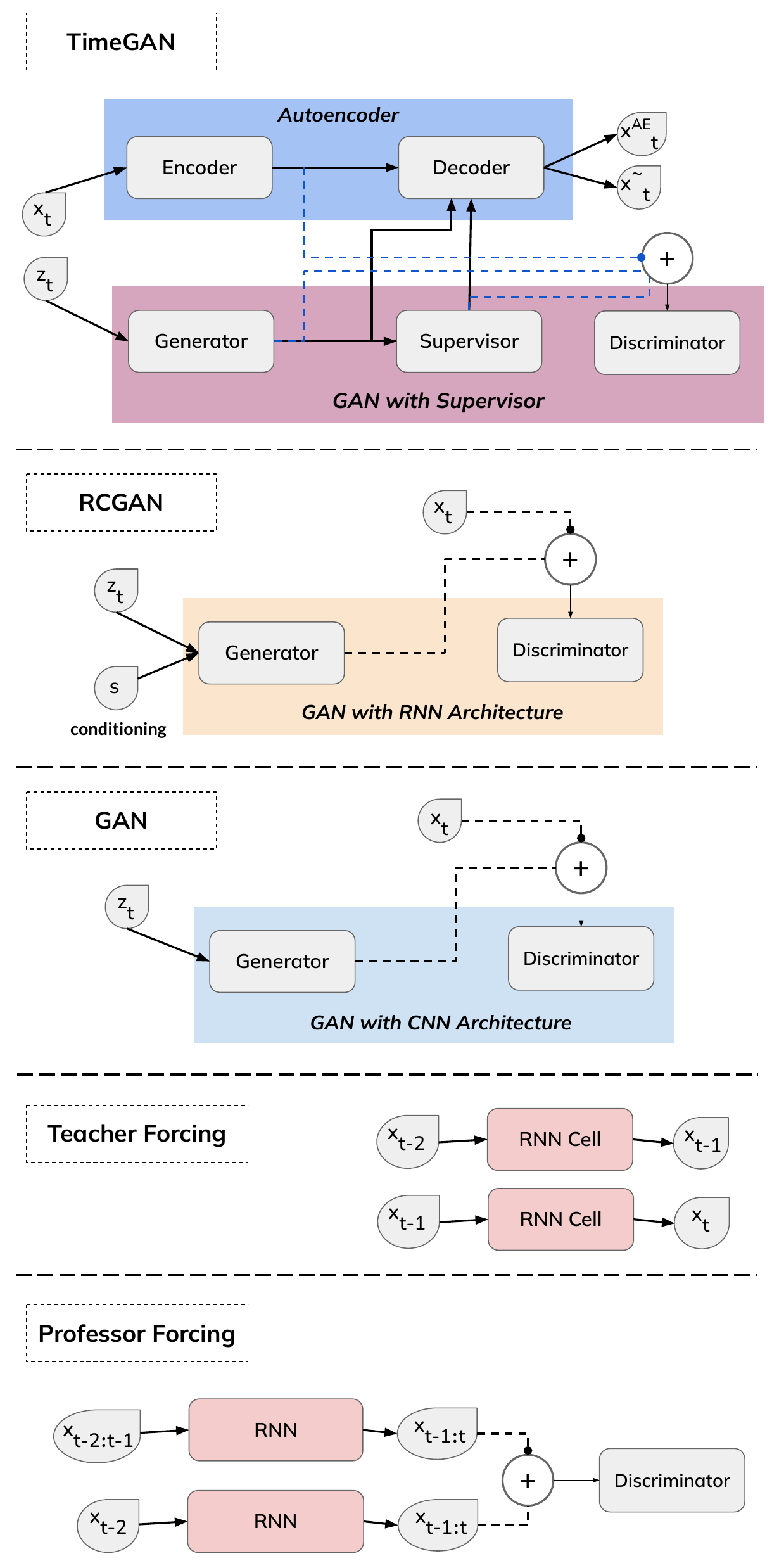}
\caption{The figure illustrates the architecture of five distinct time series generation techniques: TimeGAN, RCGAN, GAN, Teacher Forcing, and Professor Forcing. Each method presents a unique approach to generating time series data, offering various strengths and applications depending on the specific requirements of the task.}
\label{fig:baseline}
\end{figure}

The SeriesGAN framework is developed to improve the performance and robustness of time series generation methods, specifically TimeGAN, by achieving several key goals. First, it enhances data reconstruction by the decoder and data generation by the generator through the training of two discriminators in both the latent and feature spaces. Second, it facilitates the convergence of the generator network by implementing a novel autoencoder-based loss function that guides the generator by providing characteristics of real time series data. Third, it incorporates a teacher forcing supervisor along with a novel loss function, trained jointly with the generator as a combined network. This strengthens the generator’s ability to learn and capture the temporal dynamics of time series data more effectively. Fourth, it integrates LSGANs instead of standard GANs and includes an early stopping algorithm to enhance stability and achieve consistently optimal results under the same hyperparameters. Fig. \ref{fig:baseline} showcases the architecture of the described time series generation baselines, and Fig. \ref{fig:seriesgan} shows the architecture of SeriesGAN.

\section{Problem Formulation}
\label{sec:problem}

In this setting, we work with data that contains temporal features (i.e., features that evolve over time, such as sensor measurements). 
Let $\mathcal{X}$ represent the space of these temporal features, and let $\mathbf{X} \in \mathcal{X}$ be random vectors, each of which can take on specific values denoted by $\mathbf{x}$. 
We consider sequences of temporal data, denoted $\mathbf{X}_{1:T}$, drawn from a joint distribution $p$. The length $T$ of each sequence is itself a random variable, which is absorbed into the distribution $p$. In our training data, each sample is indexed by $n \in \{1, \ldots, N\}$, and the dataset can be represented as 
$\mathcal{D} = \{\mathbf{x}_{n,1:T_n}\}_{n=1}^{N}$. 
From this point forward, we omit the subscript $n$ unless needed for clarity.

Our primary goal is to use the training data $\mathcal{D}$ to estimate a probability density function $\hat{p}(\mathbf{X}_{1:T})$ that closely approximates the true distribution $p(\mathbf{X}_{1:T})$. 
This is a challenging objective, particularly given the variability in sequence lengths, the dimensionality of the data, and the complexity of the distribution. 
To address this, we decompose the joint distribution $p(\mathbf{X}_{1:T})$ autoregressively as $p(\mathbf{X}_{1:T}) = \prod_t p(\mathbf{X}_t|\mathbf{X}_{1:t-1})$. 
This allows us to focus on a simpler objective: learning a conditional density function $\hat{p}(\mathbf{X}_t|\mathbf{X}_{1:t-1})$ that approximates the true $p(\mathbf{X}_t|\mathbf{X}_{1:t-1})$ at any given time step $t$.

The two objectives are as follows:

1. \textbf{Global objective}: The first objective aims to match the entire sequence-level joint distribution between the true and estimated data. This can be formalized as:
\begin{equation}
    \min_{\hat{p}} D\left( p(\mathbf{X}_{1:T}) \middle\| \hat{p}(\mathbf{X}_{1:T}) \right),
\end{equation}
where $D$ is a suitable distance measure between the two distributions.

2. \textbf{Local objective}: The second objective focuses on matching the conditional distributions at each time step for all $t$, which can be expressed as:
\begin{equation}
    \min_{\hat{p}} D\left( p(\mathbf{X}_t|\mathbf{X}_{1:t-1}) \middle\| \hat{p}(\mathbf{X}_t|\mathbf{X}_{1:t-1}) \right),
\end{equation}

In the ideal case for a GAN framework, the global objective corresponds to minimizing the Jensen-Shannon divergence between the real and estimated distributions. On the other hand, the local objective, under supervised learning, corresponds to minimizing the Kullback-Leibler divergence. 
Minimizing the global objective assumes the presence of a perfect discriminator (which may not always be accessible), while minimizing the local objective requires access to ground-truth sequences (which are available in this case). 
Thus, our approach combines the GAN objective (related to the global distribution) with a supervised learning objective (focused on the conditionals). 
Ultimately, this results in a training process that integrates adversarial learning with teacher forcing-based autoregressive learning, guiding the model toward more precise approximations.

\section{Proposed Model}
\label{sec:method}

\begin{figure}
\centering
\includegraphics[width=0.45\textwidth]{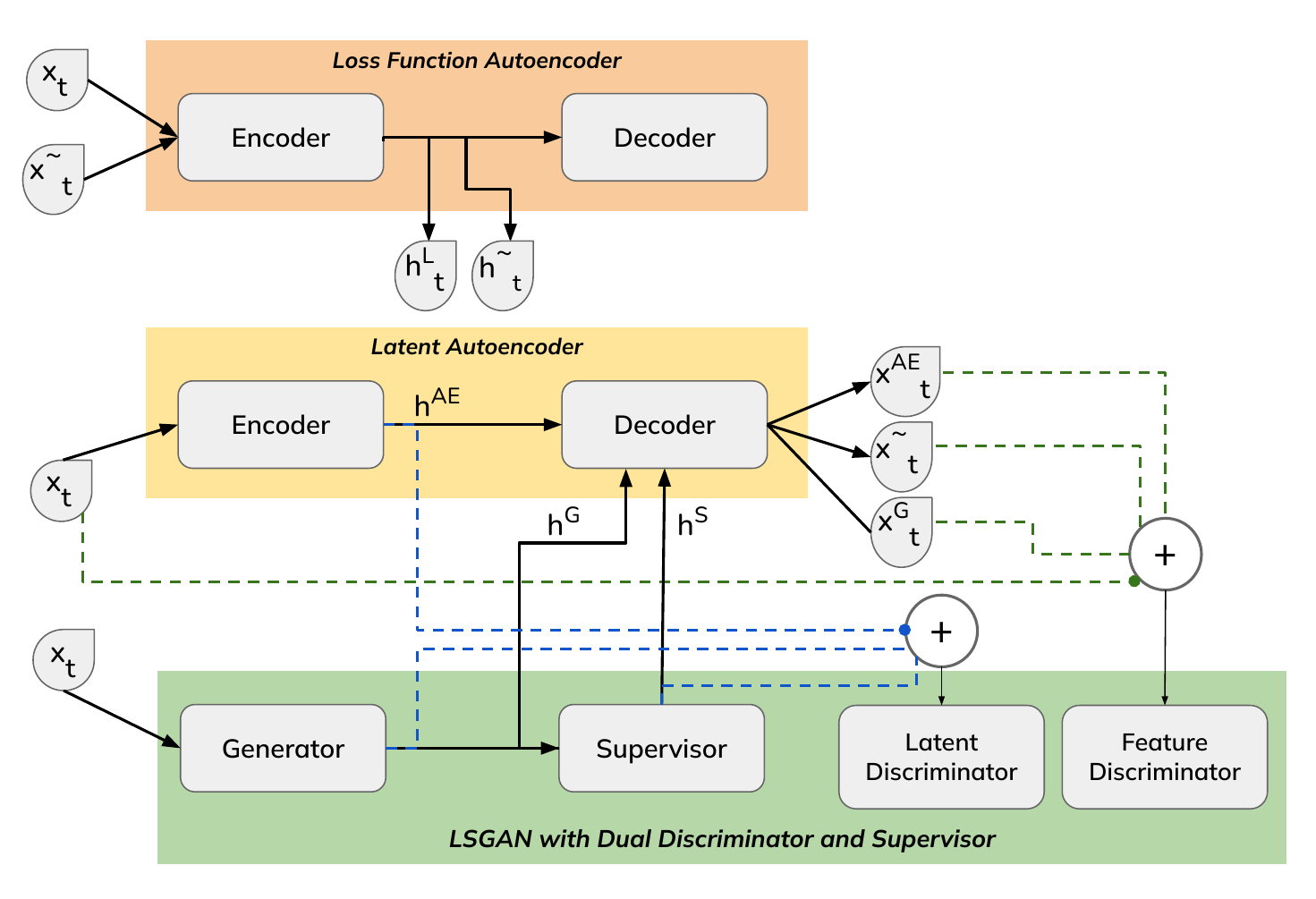}
\caption{The figure showcases the architecture of SeriesGAN for time series generation. It includes two autoencoders, which play a crucial role in loss function calculation and facilitate lower-dimensionality training of the GAN network. Additionally, it incorporates two discriminators that enhance the data reconstruction capabilities of the autoencoder and improve the data generation quality of the generator network.}
\label{fig:seriesgan}
\end{figure}

As illustrated in Fig. \ref{fig:seriesgan}, the framework consists of six networks: two autoencoders, referred to as the \textit{loss function autoencoder} and the \textit{latent autoencoder}, a generator, a supervisor, and two discriminators, named the \textit{latent discriminator} and the \textit{feature discriminator}. The loss function autoencoder is employed to implement our novel time series loss, which facilitates the generator network in more effectively capturing the intrinsic characteristics of real time series data. The latent autoencoder’s role is to facilitate training by generating compressed representations in the latent space, thereby reducing the likelihood of non-convergence within the GAN framework \cite{AGGARWAL2021100004}. The generator produces data in this lower-dimensional latent space rather than in the feature space. The supervisor network, integrated with a novel supervised loss function, is specifically designed to learn the temporal dynamics of the time series data through teacher-forcing training \cite{Tforcing}. This is crucial, as relying solely on the discriminators’ binary adversarial feedback may not sufficiently prompt the generator to capture the data’s stepwise conditional distributions. The latent discriminator provides efficient feedback to the generator by distinguishing between real and fake data in the latent space, while the feature discriminator differentiates between fake and real data in the feature space, providing secondary and more accurate feedback to both the generator and the autoencoders.

As illustrated in Fig. \ref{fig:seriesgan}, the input data $\mathbf{x}$ is encoded into the latent space $\mathbf{h}^{AE}$ using the encoder function $e$, where $\mathbf{h}^{AE} = e_{\mathcal{X}}(\mathbf{x})$. This encoding captures the essential features of $\mathbf{x}$ through the latent autoencoder. The data reconstruction, $\mathbf{x}^{AE} = r_{\mathcal{X}}(\mathbf{h}^{AE})$, is then achieved by decoding $\mathbf{h}^{AE}$ with the recovery function $r$, aiming to closely replicate the original input. The generator function $g$ transforms a random noise vector $\mathbf{z}$ into synthetic latent data $\mathbf{h}^{G} = g_{\mathcal{X}}(\mathbf{z})$, which is subsequently reconstructed into synthetic data $\mathbf{x}^{G} = r_{\mathcal{X}}(\mathbf{h}^{G})$. To further refine the synthetic data, the supervisor network $s$ processes $\mathbf{h}^{G}$ to generate a latent representation $\mathbf{h}^{S} = s_{\mathcal{X}}(\mathbf{h}^{G})$, from which the final synthetic data $\tilde{\mathbf{x}} = r_{\mathcal{X}}(\mathbf{h}^{S})$ is reconstructed. Additionally, the encoding of the input data $\mathbf{x}$ can be represented as $\mathbf{h}^{L} = \hat{e}_{\mathcal{X}}(\mathbf{x})$, where $\hat{e}$ is the encoder function of the loss function autoencoder, designed to capture the key characteristics of $\mathbf{x}$. Similarly, the encoding of the synthetic data $\tilde{\mathbf{x}}$ is obtained as $\tilde{\mathbf{h}} = \hat{e}_{\mathcal{X}}(\tilde{\mathbf{x}})$.

\subsection{Autoregressive Learning}

Combining the GAN framework with autoregressive learning enables us not only to approximate the true distribution \( p(\mathbf{X}_{1:T}) \) through a learned probability density function \( \hat{p}(\mathbf{X}_{1:T}) \), but also to model the conditional density function \( \hat{p}(\mathbf{X}_t|\mathbf{X}_{1:t-1}) \) that approximates the true \( p(\mathbf{X}_t|\mathbf{X}_{1:t-1}) \) at any given time step \( t \). To achieve this, SeriesGAN utilizes a GRU-based supervisor network trained alongside the generator to meet both objectives. 

The SeriesGAN framework consists of four distinct training phases. In the first two phases, the two autoencoders are trained independently of the other networks to effectively learn the encoding and decoding representations of the real data. This isolated training ensures that each autoencoder captures the underlying structure of the data before integrating with the rest of the model. In the third phase of training, the supervisor network is separately trained to predict the second next timestamp \( t \) by leveraging timestamps \(1\) to \( t-2 \), which leads to improved performance compared to the conventional approach of predicting the next timestamp \( t \) based on \(1\) to \( t-1 \). In the fourth phase, the generator and supervisor are trained together as a single integrated network using a combined loss function \( \mathcal{L}_G \), which updates the weights of both components. During this phase, the latent and feature autoencoders undergo joint training with the generator-supervisor model in an adversarial framework. The loss \( \mathcal{L}_G \) in \eqref{eq:three} is composed of multiple sub-losses, including the supervised loss \( \mathcal{L}_S \), which captures the temporal dynamics during teacher-forcing training, and adversarial feedback losses \( \mathcal{L}_{U_{latent}} \) and \( \mathcal{L}_{U_{feature}} \) from the latent and feature discriminators, respectively. Moreover, SeriesGAN incorporates the Mean Absolute Error (MAE) between the real data \( \mathbf{x} \) and the generated data \( \tilde{\mathbf{x}} \), represented by \( \mathcal{L}_V \), ensuring the generated data closely matches the statistical properties of the real data. The network also introduces a novel autoencoder-based time series loss \( \mathcal{L}_{TS} \), which further enhances the quality of the generated data by embedding the characteristics of the training data into the generator, leading to more accurate and reliable outputs.

\begin{equation}
\label{eq:three}
\mathcal{L}_{G} = \mathcal{L}_{U_{latent}} + \mathcal{L}_{U_{feature}} + \mathcal{L}_S + \mathcal{L}_V + \mathcal{L}_{TS}
\end{equation}

\begin{equation}
\mathcal{L}_V = \mathcal{L}_{Mean} + \mathcal{L}_{Variance}
\end{equation}

The sub-loss function $\mathcal{L}_V$ is designed to assist the generator in learning the distribution of the real data by providing the differences between the mean and variance of a batch of real and synthetic data.

\begin{equation}
\mathcal{L}_{\text{Mean}} = \mathbb{E}_{x_{1:T} \sim p} \left[ \sum_t \left| \frac{1}{N} \sum_{n=1}^N{\mathbf{x}}_{t_n} - \frac{1}{N} \sum_{n=1}^N  {\tilde{\mathbf{x}}_{t_n}} \right| \right]
\end{equation}

Where $\mathcal{L}_{\text{Mean}}$ computes the MAE between the mean of a batch of real data $\mathbf{x}$ and synthetic data $\tilde{\mathbf{x}}$. We consider sequences of temporal data, denoted $\mathbf{X}_{1:T}$, drawn from a joint distribution $p$, where each sample is indexed by $n \in \{1, \ldots, N\}$ and the batch is represented as $\mathcal{B} = \{{\mathbf{x}}_{n,1:T_n}\}_{n=1}^{N}$.

\begin{align}
\mathcal{L}_{\text{Variance}} = & \, \mathbb{E}_{x_{1:T} \sim p} \left[ \sum_t \left| \frac{1}{N} \sum_{n=1}^N ( {\mathbf{x}}_{t_n} - \overline{{\mathbf{x}_t}})^2 \right. \right. \nonumber \\
& \left. \left. - \frac{1}{N} \sum_{n=1}^N({\tilde{\mathbf{x}}}_{t_n} - \overline{{\tilde{\mathbf{x}}_{t}}})^2 \right| \right]
\end{align}

Where $\mathcal{L}_{\text{Variance}}$ measures the MAE between the variance of a batch of real data $\mathbf{x}$ and synthetic data $\tilde{\mathbf{x}}$. Here, $\overline{{\mathbf{x}}}$ represents the mean of ${\mathbf{x}}$, and $\overline{{\tilde{\mathbf{x}}}}$ represents the mean of $\tilde{{\mathbf{x}}}$ for a batch of data.

Our novel contribution, \( \mathcal{L}_S \), enhances the learning process by introducing additional structure. In closed-loop mode, the generator receives sequences of real data embeddings \( {\mathbf{h}}_{1:t-2} \) (produced by the latent autoencoder) to predict the latent vector at the second next time step \( {\mathbf{h}}_{t} \) . Gradients are calculated from a loss that measures the divergence between the distributions \( p({\mathbf{H}}_t|\mathbf{H}_{1:t-2}) \) and \( \hat{p}({\mathbf{H}}_t|\mathbf{H}_{1:t-2}) \). Using maximum likelihood estimation, this leads to the well-known supervised loss,

\begin{equation}
\mathcal{L}_S = \mathbb{E}_{x_{1:T}\sim p} \left[ \sum_t \left\| \mathbf{h}^G_t - s_{\mathcal{X}}(\mathbf{h}^G_{t-2}) \right\|_2 \right]
\end{equation}

Where \( s \) represent the supervisor network function, while \( \mathbf{h}_t^G \) denotes the output of the generator at timestamp \( t \).

\subsection{Dual-Discriminator Training}

Training a GAN framework alongside an autoencoder enables the generator to produce data within the latent space, which has lower dimensionality compared to the feature space. This reduces the likelihood of non-convergence, a common issue in GAN networks, particularly when handling high-dimensional multivariate time series data. However, adversarial training within the latent space can lead to the generation of noisy data for two key reasons. First, the generator relies on adversarial feedback in the latent space, which, while more efficient, is often insufficient since some data characteristics are inevitably lost during encoding. Second, the autoencoder struggles to fully preserve the generated data’s attributes when only reconstruction loss is considered \cite{blurred}. On the other hand, discrimination within the feature space provides more accurate, though less efficient, feedback for the generator. This type of feedback can also be beneficial to the autoencoder, aiding it in improving its decoding performance through adversarial feedback mechanisms. To leverage the benefits of both latent and feature space feedback, the SeriesGAN framework incorporates two discriminators, ensuring the advantages of both spaces are utilized.

Through a joint learning scheme, the loss function autoencoder is initially trained using only the reconstruction loss. This trained network is then leveraged in the fourth phase, where the \(\mathcal{L}_{TS}\) loss is applied to the generator network to further refine the quality of the generated data. In the second phase, the latent autoencoder is trained using a combination of reconstruction loss \(\mathcal{L}_{R}\) and binary feedback from the feature discriminator \(\mathcal{L}_{U_{joint}}\) , where real data is the dataset (\(\mathbf{x}\)) and fake data is its reconstruction (\(\mathbf{x}^{AE}\)), as shown in \eqref{eq:eight} and \eqref{eq:nine}. This approach enhances the latent autoencoder's precision in reconstructing outputs. In the third phase, the supervisor network is trained using supervised loss as discussed earlier. In the fourth phase, all five networks (except for the loss function autoencoder) are trained jointly. During this phase, the feature discriminator distinguishes between real data (the dataset, \(\mathbf{x}\)) and the dataset reconstructions (\(\mathbf{x}^{AE}\)), with the fake data comprising the generator's decoded outputs (\(\mathbf{x}^{G}\)) and the supervisor’s decoded outputs (\(\tilde{\mathbf{x}}\)). The latent discriminator differentiates between the encoder's output (\(\mathbf{h}^{AE}\)) as real data and the outputs of the generator (\(\mathbf{h}^{G}\)) and the supervisor (\(\mathbf{h}^{S}\)) as fake data. The generator and supervisor networks undergo integrated training through these two adversarial feedback mechanisms, along with other feedback mechanisms including \(\mathcal{L}_S\), \(\mathcal{L}_V\), and \(\mathcal{L}_{TS}\). This fourth phase involves a shift in the characterization of fake and real data for the feature discriminator compared to the second phase.

\begin{equation}
\label{eq:eight}
\mathcal{L}_{AE} = \mathcal{L}_{R} + \mathcal{L}_{U_{joint}} \textbf{;} \quad
\mathcal{L}_{R} = \mathbb{E}_{x_{1:T}\sim p} \left[ \sum_t \|{\mathbf{x}}_t - {{\mathbf{x}}^{AE}}_t\|_2 \right]
\end{equation}

$\mathcal{L}_{R}$ refers to the conventional reconstruction loss, which is used for training autoencoders.

\begin{equation}
\label{eq:nine}
\mathcal{L}_{U_{joint}}  = \mathbb{E}_{x_{1:T}\sim p} \left[ \sum_t \log y_t \right] + \mathbb{E}_{x_{1:T}\sim \hat{p}} \left[ \sum_t \log (1 - \tilde{y}_t) \right]
\end{equation}

Let \(\tilde{y} = d_{\mathcal{X}}(\mathbf{x}^{AE})\) and \(y = d_{\mathcal{X}}(\mathbf{x})\), where \(d\) represents the feature discriminator function. The probability distribution of the real data is denoted by \(p\), while \(\hat{p}\) corresponds to the probability distribution of the synthetic data.

\subsection{Autoencoder-based Loss Function}

We introduce a novel loss function to better guide the generator network in learning the characteristics of the dataset. Providing only adversarial feedback to the generator is insufficient for teaching the generator the nuances of time series characteristics, resulting in synthetic data that does not closely resemble real data. To address this issue, it is essential to supply the generator with the intrinsic properties of the dataset. However, this task is challenging due to the numerous features present in time series data, including trend, seasonality, and cyclicity \cite{10020669}. By employing a GRU autoencoder, named the loss function autoencoder, and training it with reconstruction loss on the dataset, we can extract compressed features of the time series samples via the encoder \cite{7965877}. We then calculate the loss function as the mean squared error (MSE) of the mean and standard deviation (std) between the compressed versions of a batch of real (\(\mathbf{h}^{L}\)) and synthetic (\(\tilde{\mathbf{h}}\)) data. This loss is termed \(\mathcal{L}_{TS}\). Equations \eqref{eq:ten}, \eqref{eq:eleven}, and \eqref{eq:twelve} provide the mathematical formulation of this loss function. The mapping function $\hat{e}$ serves as the encoder for this autoencoder. We train the loss function autoencoder prior to training the overall framework (first phase). Unlike the latent autoencoder, which compresses the attribute dimension of a multivariate time series (MVTS), this particular autoencoder compresses the timestamp dimension of an MVTS.

\begin{equation}
\label{eq:ten}
\mathcal{L}_{TS} = \mathcal{L}_{TS_{mean}} + \mathcal{L}_{TS_{std}}
\textbf{;} \quad
\mathbf{h}^{L} = \hat{e}_{\mathcal{X}}(\mathbf{x})
\textbf{;} \quad
\tilde{\mathbf{h}} = \hat{e}_{\mathcal{X}}(\tilde{\mathbf{x}})
\end{equation}

where $\hat{e}$ is the encoder of the loss function autoencoder, $\mathbf{x}$ represents the real data, and $\tilde{\mathbf{x}}$ denotes the synthetic data.

\begin{equation}
\label{eq:eleven}
\mathcal{L}_{TS_{mean}} = \mathbb{E}_{x_{1:T} \sim p} \left[ \sum_t  \left\| \frac{1}{N} \sum_{n=1}^N {\mathbf{h}^{L}_{t_n}} - \frac{1}{N} \sum_{n=1}^N {\tilde{\mathbf{h}}_{t_n}} \right\|_2 \right]
\end{equation}

$\mathcal{L}_{TS_{mean}}$ calculates the MSE between the mean values of a batch of $\mathbf{h}^{L}$ and $\tilde{\mathbf{h}}$.

\begin{align}
\label{eq:twelve}
\mathcal{L}_{TS_{std}} = & \mathbb{E}_{x_{1:T} \sim p} \Bigg[ \sum_t \Bigg\| \sqrt{\frac{1}{N} \sum_{n=1}^N ({\mathbf{h}^{L}_{t_n}} - \overline{{\mathbf{h}^{L}_t}})^2} \nonumber \\
& - \sqrt{\frac{1}{N} \sum_{n=1}^N ({\tilde{\mathbf{h}}_{t_n}} - \overline{{\tilde{\mathbf{h}}_t}})^2} \Bigg\|_2 \Bigg]
\end{align}

where $\mathcal{L}_{TS_{std}}$ calculates the MSE between the std values of a batch of $\mathbf{h}^{L}$ and $\tilde{\mathbf{h}}$.

\subsection{Early Stopping and LSGANs}

Another significant issue with GANs is stability, and TimeGAN is not exempt from this challenge. To improve the framework's stability, we implement an early stopping algorithm, acknowledging that the best results may occur after a random rather than a fixed number of iterations. As outlined in Algorithm \ref{alg:earlystopping}, after completing half of the total epochs, synthetic data generation begins, and the discriminative score between real and synthetic data is calculated every 500 epochs. Additionally, we calculate the MSE of both the mean and std between the embeddings of the real and synthetic data, which are obtained via the loss function autoencoder. This is the same process as calculating \( \mathcal{L}_{TS} \), to evaluate how closely the synthetic data matches the real data in terms of its characteristics. By combining the discriminative score with \( \mathcal{L}_{TS} \), we decide whether to save the current model and the generated data. At the end of the training process, we ensure the framework delivers optimal saved results, consistently producing reliable and quality outcomes. Determining the correct weighting of the two metrics is essential to effectively compare the current model with previously saved versions. The balance between the discriminative score and \( \mathcal{L}_{TS} \) can vary depending on the dataset, making fixed hyperparameters impractical. To handle this, the hyperparameter \( p_1 \) is calculated during the first evaluation of these metrics and is then used consistently in subsequent epochs.

Furthermore, we employ LSGANs \cite{mao2017squares} instead of standard GANs as they provide more stable training.  More specifically, standard GANs utilize a binary cross-entropy loss, which can lead to issues with vanishing gradients. LSGANs, on the other hand, use a least squares loss function, also known as MSE loss, that mitigates this problem by providing smoother gradients. This difference results in more stable and effective training dynamics for LSGANs compared to standard GANs.

\begin{algorithm}
\caption{Early Stopping Algorithm}
\label{alg:earlystopping}
\small
\begin{algorithmic}
\STATE Initialize $real$ and $synthetic$ samples
\STATE Initialize $\mathbf{h}^L$ and $\tilde{\mathbf{h}}$
\STATE Set $N$ as the total number of epochs
\STATE Initialize $totalError$ and $p1$ to None
\STATE Set $checkEpoch \gets 500$ and $startEpoch \gets \lfloor \frac{N}{2} \rfloor$
\STATE
\FOR {$epoch = 1$ \TO $N$}
    \IF {$epoch \geq startEpoch$ \AND $epoch \bmod checkEpoch == 0$}
        \STATE $disScore \gets \text{calcDiscriminate}(real, synthetic)$
        \STATE $meanReal \gets \text{calcMean}(\mathbf{h}^L)$
        \STATE $meanSynth \gets \text{calcMean}(\tilde{\mathbf{h}})$
        \STATE $mseMean \gets \text{calcMSE}(meanReal, meanSynth)$
        
        \STATE $varReal \gets \text{calcVar}(\mathbf{h}^L)$
        \STATE $varSynth \gets \text{calcVar}(\tilde{\mathbf{h}})$
        \STATE $mseVar \gets \text{calcMSE}(varReal, varSynth)$
        \STATE $mseSTD \gets \sqrt{mseVar}$

        \IF {$p1 == \text{None}$}
            \STATE $p1 \gets \frac{disScore}{mseMean + mseSTD}$
        
        \ENDIF
        \STATE $score \gets disScore + p1 * (mseMean + mseSTD)$
        
        \IF {$score \leq totalError$ \OR $totalError == \text{None}$}
            \STATE $totalError \gets score$
            \STATE $\text{saveSynthetic}(synthetic)$
        \ENDIF
    \ENDIF
\ENDFOR
\end{algorithmic}
\end{algorithm}

\section{Experiments}
\label{sec:experiments}

\subsection{Code Repository and Hyperparameters}
The SeriesGAN's codebase, along with an extensive tutorial on its usage, implementation specifics, and hyperparameters, is accessible to the public for review and application \footnote{The codebase of SeriesGAN is available here: \href{https://github.com/samresume/SeriesGAN}{https://github.com/samresume/SeriesGAN}}.

We have designed SeriesGAN in such a way that by simply calling a Python function and passing the time series data along with the desired hyperparameters, the network initiates training and, once complete, generates as many synthetic samples as needed.

\subsection{Datasets}
We assess SeriesGAN's performance on time series datasets that exhibit diverse characteristics, including periodicity, noise levels, length, and feature correlation. The datasets are chosen based on various combinations of these characteristics.

\begin{enumerate}
    
    \item \textbf{Medical (ECG):} The ECG5000 dataset from Physionet, consisting of a 20-hour ECG recording with 140 time points per sample, represents a univariate time series that is both continuous and periodic. This dataset is valuable due to the similarity in temporal dynamics across different samples.

    \item \textbf{Space Weather (SWAN-SF):} The Space Weather Analytics for Solar Flares (SWAN-SF) \cite{angryk2020multivariate} dataset classifies solar flares into five categories: GOES X, M, C, B, and FQ, with the FQ class encompassing both flare-quiet and GOES A-class events. For our study, we exclusively focus on the major flare categories, GOES X and M. We utilize only the first partition out of five, which maintains an approximately equal distribution of these major flares associated with a specific active region (AR). This partition contains a multivariate time series dataset with 24 unique attributes and a sequence length of 60. This dataset is selected for its high dimensional feature space and its distinctive temporal dynamics, which include noise and variability.

    \item \textbf{Finance (Stocks):} Stock price sequences are continuous but aperiodic and features are correlated. We use daily historical data from Google stocks spanning 2004 to 2019, which includes features such as volume, high, low, opening, closing, and adjusted closing prices. This dataset exhibits characteristics such as noise, underlying trends, and randomness.

    \item \textbf{Sines:} We construct multivariate sinusoidal time series with distinct frequencies $\eta$ and phases $\theta$, resulting in continuous, periodic, and multivariate signals where each feature operates independently. For each dimension $i$ ranging from 1 to 5, the corresponding function is given by $x_i(t) = \sin(2\pi\eta t + \theta)$, where $\eta$ is sampled from a uniform distribution $U[0, 1]$ and $\theta$ is drawn from $U[-\pi, \pi]$. This dataset is valuable for analysis because its cyclic nature presents a significant challenge for any learning technique.
    
\end{enumerate}

\subsection{Evaluation Metrics and Baselines}

We perform a comparative analysis of five leading time series generation techniques: SeriesGAN, TimeGAN \cite{NEURIPS2019_c9efe5f2}, Teacher Forcing (T-Forcing) \cite{Tforcing}, Professor Forcing (P-Forcing) \cite{Pforcing}, and Standard GAN \cite{goodfellow2014generative}. These methods encompass both GAN-based and autoregressive approaches, which also form the foundation of SeriesGAN. The details of these baselines were discussed earlier in Section \ref{sec:relatedwork}. To ensure a fair comparison, identical hyperparameters, such as the type of sequence-to-sequence model and the number of layers, are consistently applied across all models.

Assessing the performance of GANs presents inherent challenges. Likelihood-based methods, such as Parzen window estimates \cite{ganevaluation1}, can produce misleading results, while the generator and discriminator losses do not directly correlate with ‘visual quality’ \cite{ganevaluation1, ganevaluation2}. Although human evaluation is often considered the most reliable approach to assessing quality, it is both impractical and costly. Furthermore, in the case of real-valued sequential data, visual inspection may not always be feasible or effective. For example, the ECG signals studied in this paper may appear random to individuals without medical expertise. Therefore, in this work, the evaluation is conducted based on three primary criteria, encompassing both qualitative and quantitative measures of the generated data.

\begin{table*}
\centering
\caption{Comparative analysis of \textbf{discriminative score} (with lower scores indicating better performance)}
\label{tbl:discriminative}
\begin{tabular}{ccccc}
\hline
\textbf{} & \textbf{Stocks} & \textbf{Sines} & \textbf{ECG} & \textbf{SWAN-SF} \\ \hline

\textbf{SeriesGAN} & \textbf{0.1873 ± 0.0823} & \textbf{0.2083 ± 0.0869} & \textbf{0.1691 ± 0.0234} & \textbf{0.2644 ± 0.0501} \\ 

TimeGAN & 0.3262 ± 0.0389 & 0.2836 ± 0.1343 & 0.2716 ± 0.0874 & 0.3745 ± 0.1014 \\ 

GAN & 0.4998 ± 0.003 & 0.3209 ± 0.2274 & 0.4863 ± 0.0174 & 0.4979 ± 0.0001 \\ 

T-Forcing & 0.4764 ± 0.0142 & 0.3482 ± 0.1358 & 0.3511 ± 0.1011 & 0.4682 ± 0.0031 \\ 

P-Forcing & 0.4832 ± 0.0021 & 0.4918 ± 0.0013 & 0.3296 ± 0.106 & 0.4964 ± 0.0017 \\ \hline

\end{tabular}
\end{table*}

\begin{table*}
\centering
\caption{Comparative evaluation of \textbf{predictive scores} (with lower scores being better)}
\label{tbl:predictive}
\begin{tabular}{ccccc}
\hline

\textbf{} & \textbf{Stocks} & \textbf{Sines} & \textbf{ECG} & \textbf{SWAN-SF} \\ \hline

\textbf{SeriesGAN} & \textbf{0.041 ± 0.0002} & \textbf{0.2232 ± 0.0018} & \textbf{0.1268 ± 0.0007} & \textbf{0.0564 ± 0.0121} \\

TimeGAN & 0.0468 ± 0.0012 & 0.2452 ± 0.001 & 0.1297 ± 0.004 & 0.0824 ± 0.0117 \\ 

GAN & 0.186 ± 0.016 & 0.2334 ± 0.0109 & 0.1916 ± 0.000 & 0.2197 ± 0.0076 \\ 

T-Forcing & 0.0501 ± 0.0011 & 0.2755 ± 0.0052 & 0.1303 ± 0.0009 & 0.066 ± 0.0107 \\ 

P-Forcing & 0.1476 ± 0.0253 & 0.2247 ± 0.0072 & 0.1942 ± 0.002 & 0.2419 ± 0.0171 \\ \hline

\end{tabular}
\end{table*}

\begin{enumerate}
  \item \textbf{Visualization}: We utilize t-SNE \cite{tsne} and PCA \cite{bryant1995principal} analyses on both the original and synthetic datasets, by flattening the temporal dimension for visualization purposes. This means that we turn the multivariate time series data into a vector by flattening the data. This approach aids in qualitatively assessing how closely the distribution of the generated samples matches that of the original in a two-dimensional space. This metric reflects one of the key characteristics of a GAN network, which aims to estimate a probability density function, \(\hat{p}(\mathbf{X}_{1:T})\), that closely approximates the true distribution \(p(\mathbf{X}_{1:T})\). It serves as a qualitative evaluation tool for assessing the performance of a generative model.
  
  \item \textbf{Discriminative Score}: For a quantitative measure of similarity, we train a post-hoc time series classification model using an LSTM to distinguish between sequences from the original and generated datasets. Each sequence from the original dataset is labeled as `real`, while each from the generated set is labeled as `synthetic`. An LSTM classifier is then trained to differentiate these two categories in a standard supervised learning task. The classification error on a reserved test set provides a quantitative measure of this score. We then subtract the result from 0.5, making the optimal result 0 instead of 0.5. This metric highlights the performance of the generated data in a downstream classification task, providing insight into how well the generated data supports real-world classification applications.
  
  \item \textbf{Predictive Score}: To assess the utility of the sampled data, it is essential that it preserves the predictive qualities of the original data. Specifically, we expect SeriesGAN to effectively capture conditional distributions over time. To evaluate this, we train an LSTM model on the synthetic dataset for sequence prediction, focusing on forecasting the next-step temporal vectors for each input sequence. The model’s accuracy is then tested on the original dataset, with performance measured using the MAE. This metric reflects the quality of the synthetic data in downstream prediction tasks, one of the most common objectives of time series analysis.
\end{enumerate}

\subsection{Evaluation Metrics Results}
For each discriminative and predictive score experiment, we conducted eight independent replications to mitigate the risk of incidental results. We utilized the GRU architecture for SeriesGAN as well as for all the baseline models. We also applied the exact same hyperparameters across all techniques, including the number of epochs, GRU layers, and batch size, ensuring a fair and consistent comparison. The mean and std for each experiment are presented in Tables \ref{tbl:discriminative} and \ref{tbl:predictive}. Based on these results, the SeriesGAN framework outperforms state-of-the-art models, including TimeGAN, Teacher Forcing, Professor Forcing, and Standard GAN, in both discriminative and predictive scores. While TimeGAN may produce variable outcomes across different training sessions, SeriesGAN consistently achieves optimal performance. In terms of the discriminative score, SeriesGAN reduces the metric by 34\% compared to TimeGAN, the current state-of-the-art. Furthermore, for the predictive score, SeriesGAN reduces the error by 12\% compared to TimeGAN. This highlights the effectiveness of the additional components integrated into SeriesGAN, which enhance both stability and overall performance. Both SeriesGAN and TimeGAN demonstrate significantly enhanced performance over the standard GAN, as shown in Tables \ref{tbl:discriminative} and \ref{tbl:predictive}. SeriesGAN achieves an average discriminative score that is 54.1\% lower and an average predictive score that is 46.2\% lower than that of the standard GAN, while TimeGAN shows a 30.4\% improvement in the discriminative score and a 39.3\% improvement in the predictive score compared to the standard GAN. This underscores the critical importance of designing GAN models specifically tailored for time series data and integrating specialized modules within standard GAN architecture to optimize it for time series generation.

\begin{figure}
\centering

\subfloat{\includegraphics[width=0.122\textwidth]{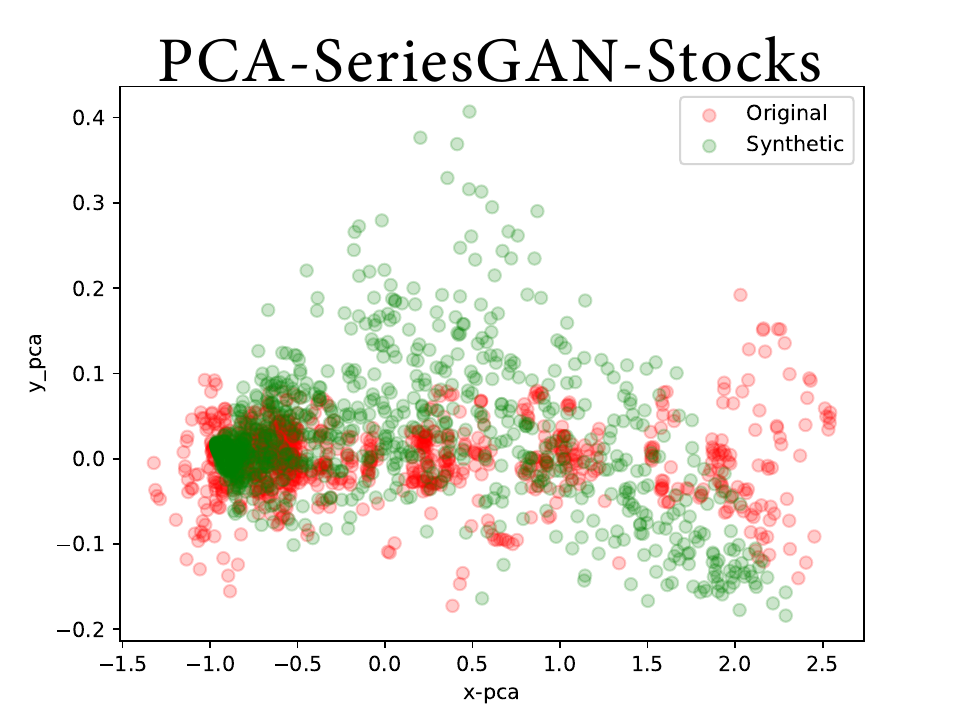}}\hfill
\subfloat{\includegraphics[width=0.122\textwidth]{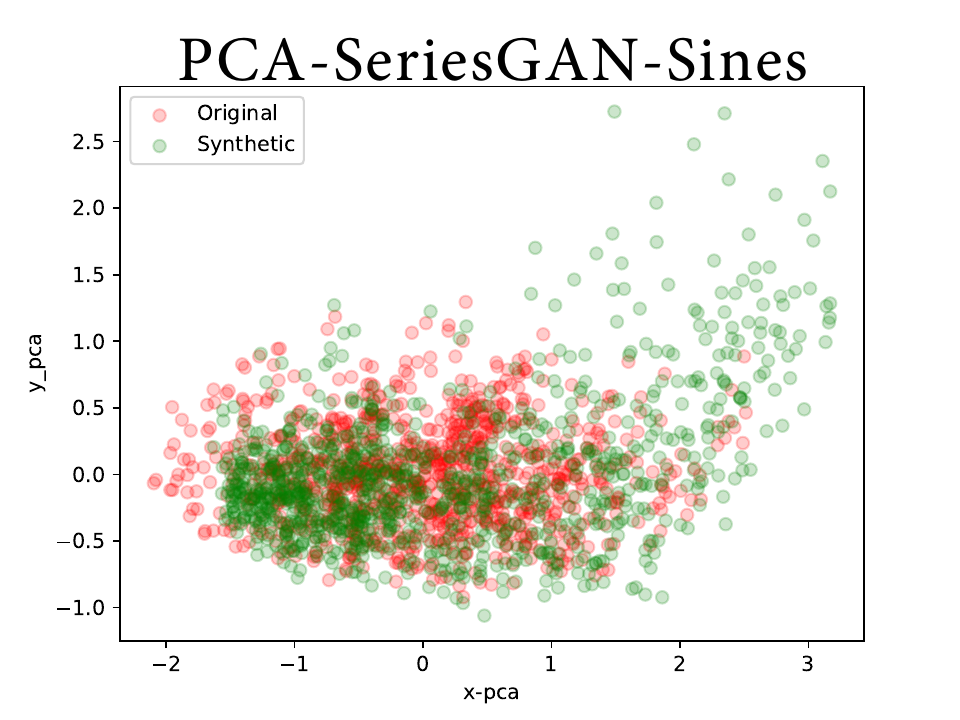}}\hfill
\subfloat{\includegraphics[width=0.122\textwidth]{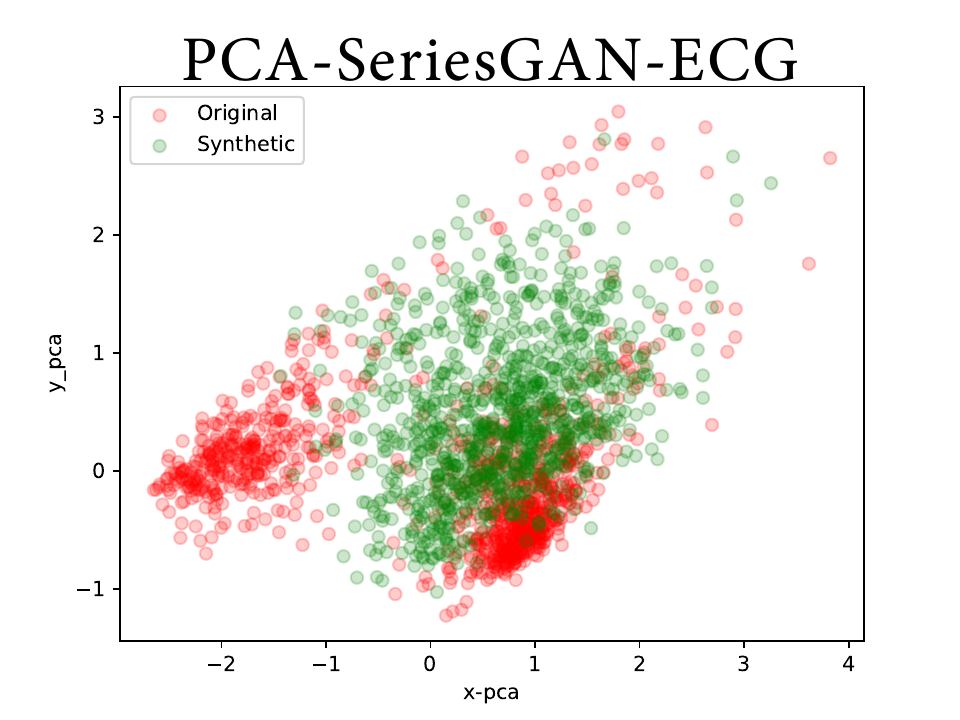}}\hfill
\subfloat{\includegraphics[width=0.122\textwidth]{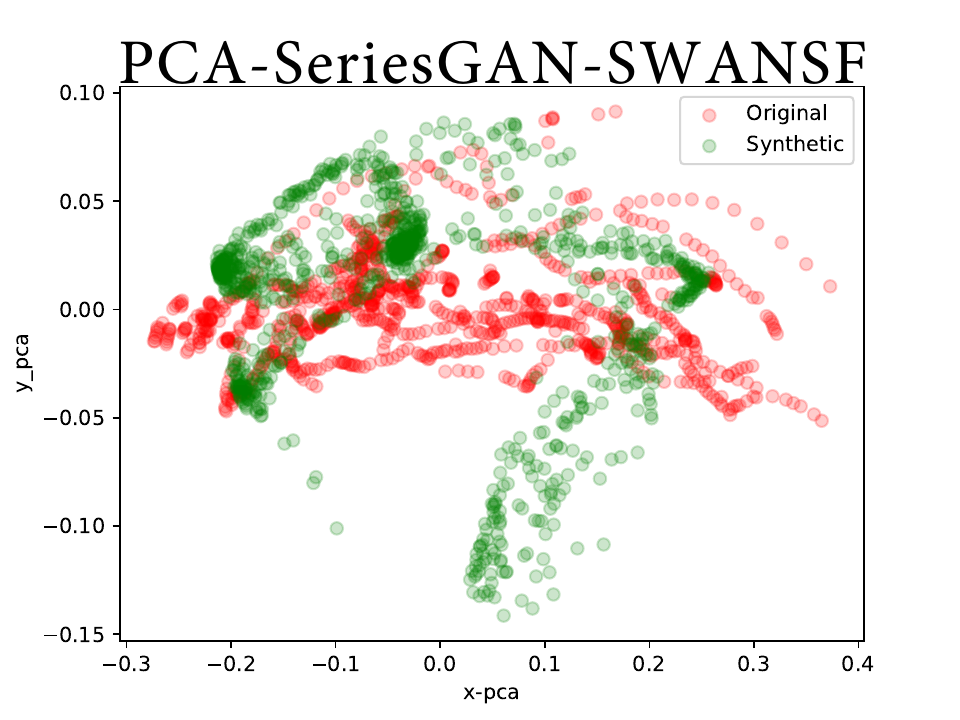}}\hfill

\vspace{-0.2cm}

\subfloat{\includegraphics[width=0.122\textwidth]{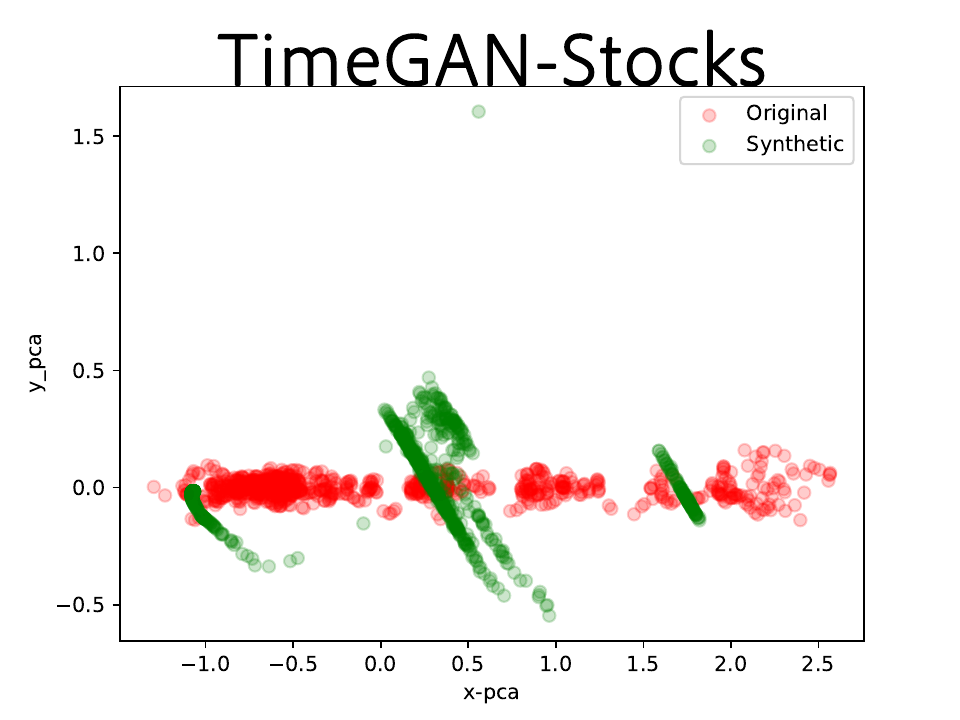}}\hfill
\subfloat{\includegraphics[width=0.122\textwidth]{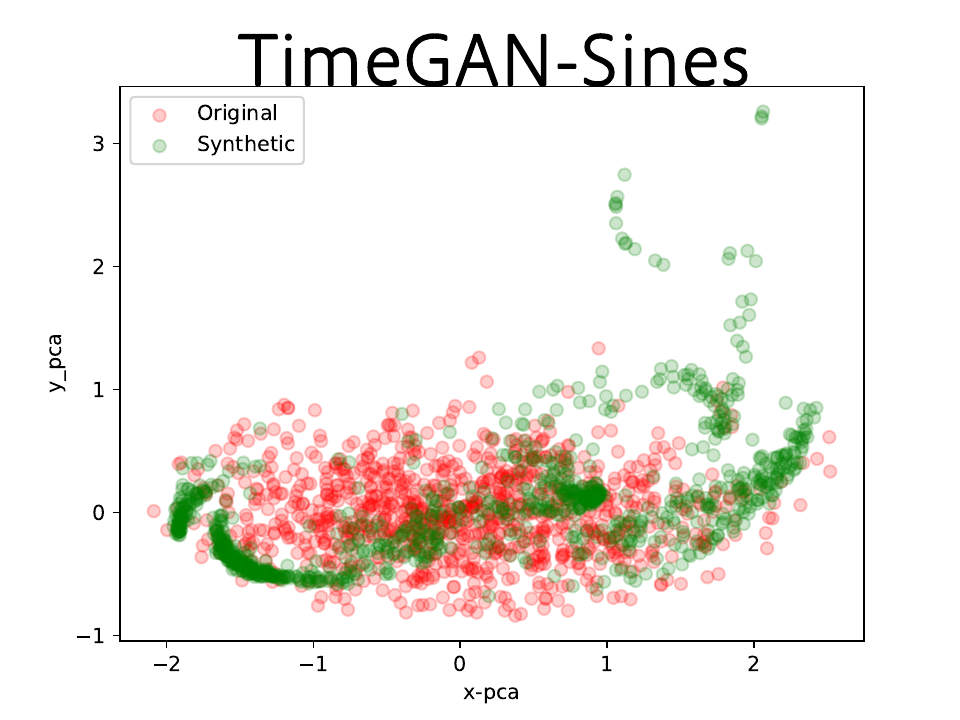}}\hfill
\subfloat{\includegraphics[width=0.122\textwidth]{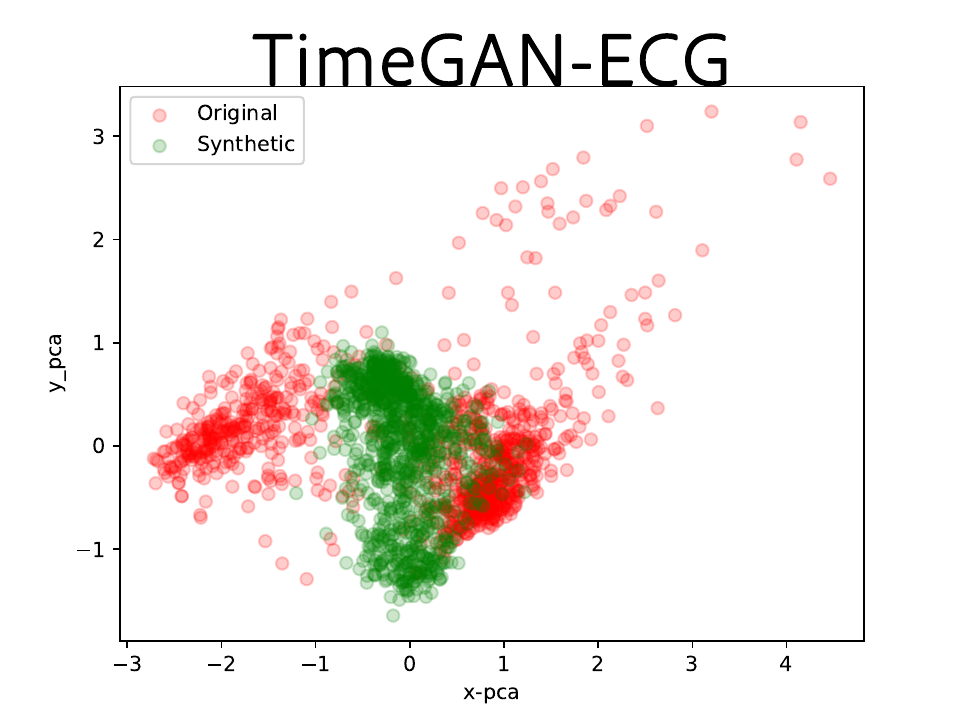}}\hfill
\subfloat{\includegraphics[width=0.122\textwidth]{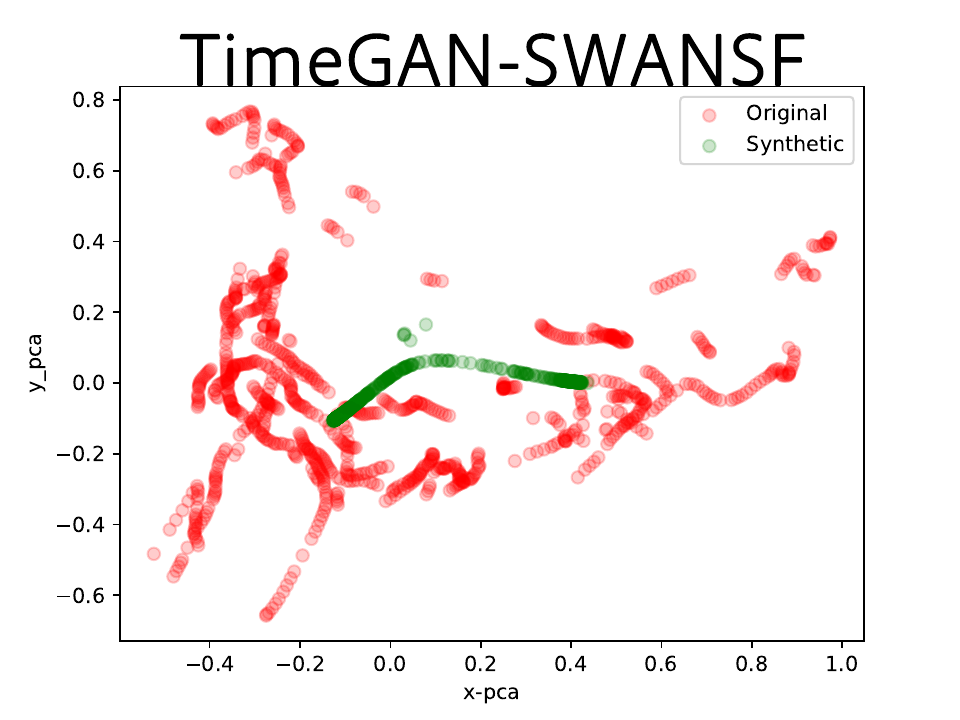}}\hfill

\vspace{-0.2cm}

\subfloat{\includegraphics[width=0.122\textwidth]{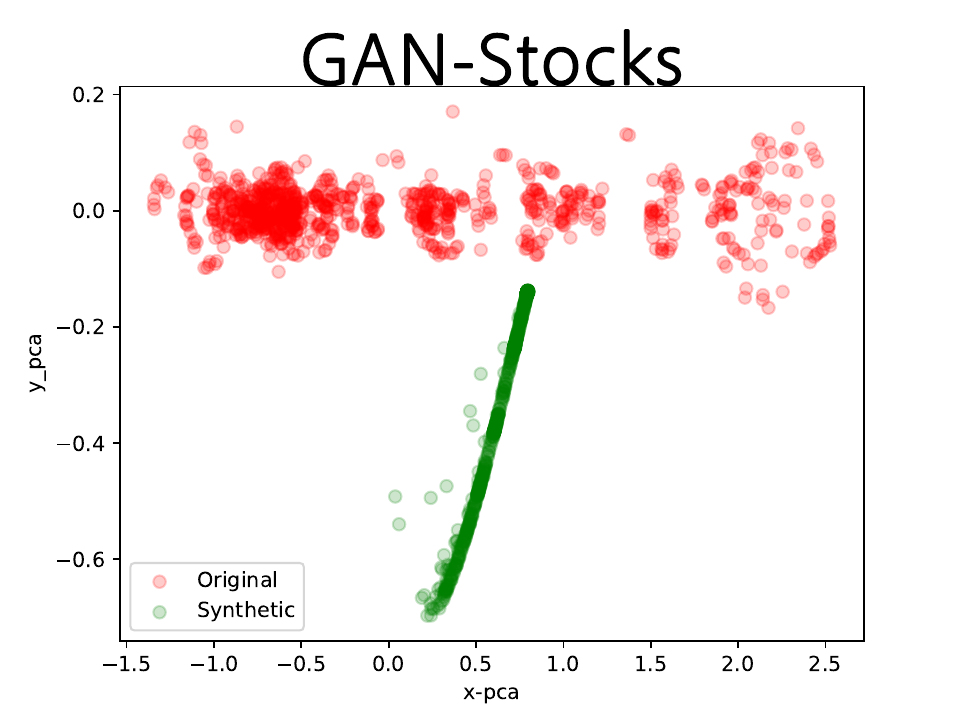}}\hfill
\subfloat{\includegraphics[width=0.122\textwidth]{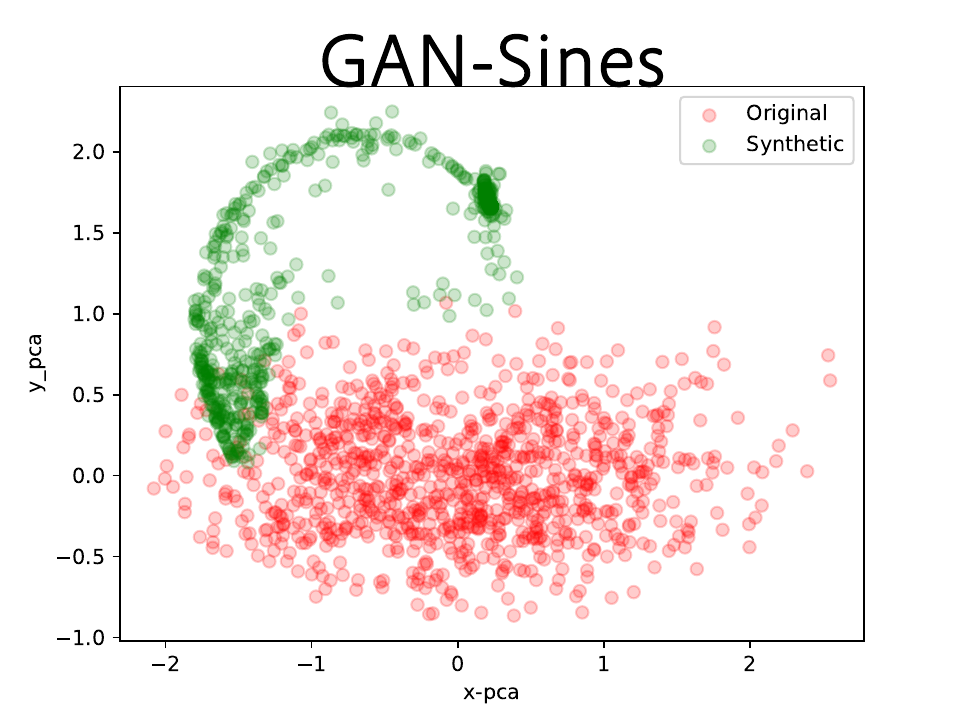}}\hfill
\subfloat{\includegraphics[width=0.122\textwidth]{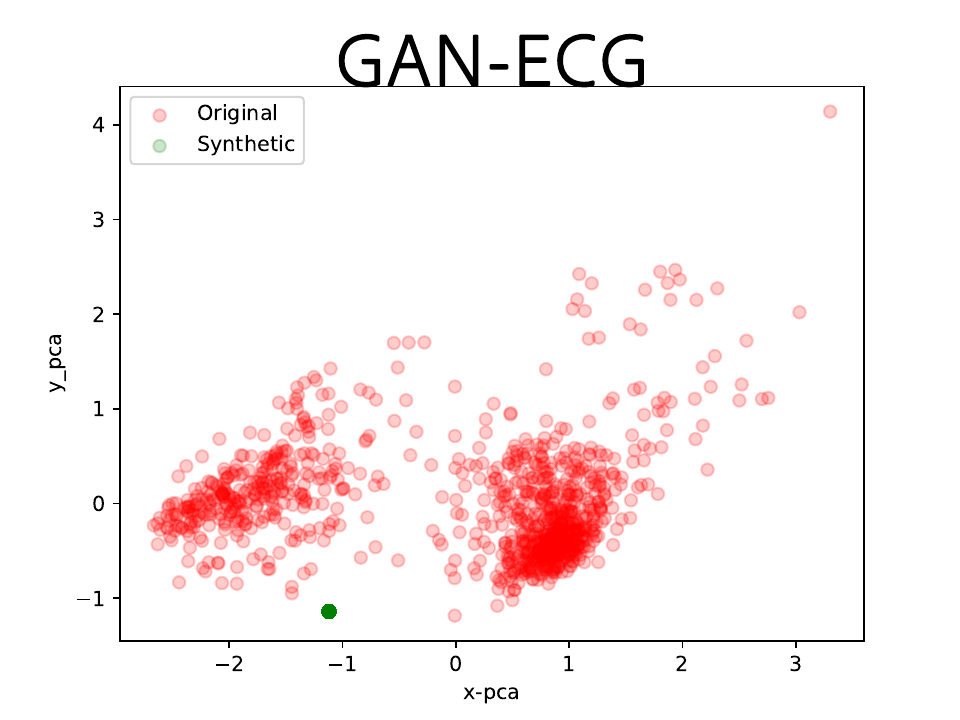}}\hfill
\subfloat{\includegraphics[width=0.122\textwidth]{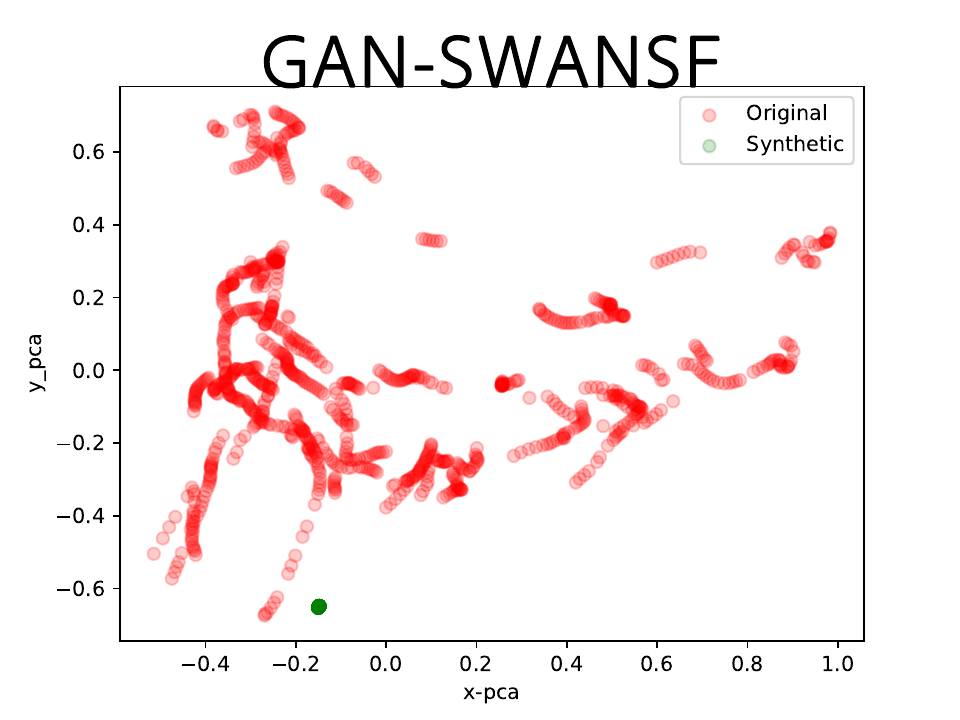}}\hfill

\vspace{-0.2cm}

\subfloat{\includegraphics[width=0.122\textwidth]{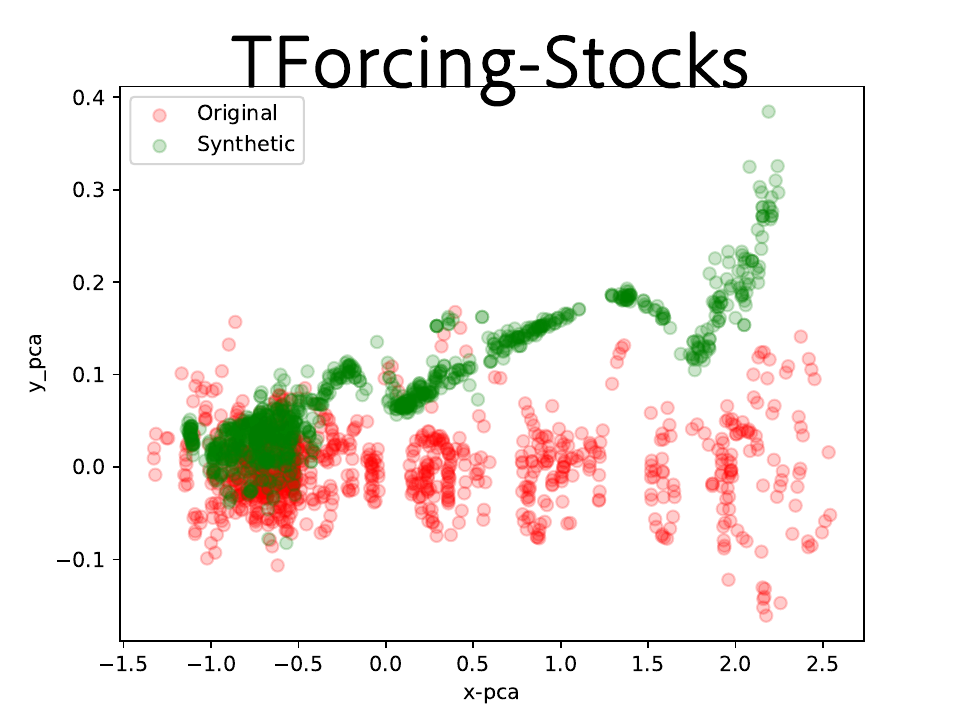}}\hfill
\subfloat{\includegraphics[width=0.122\textwidth]{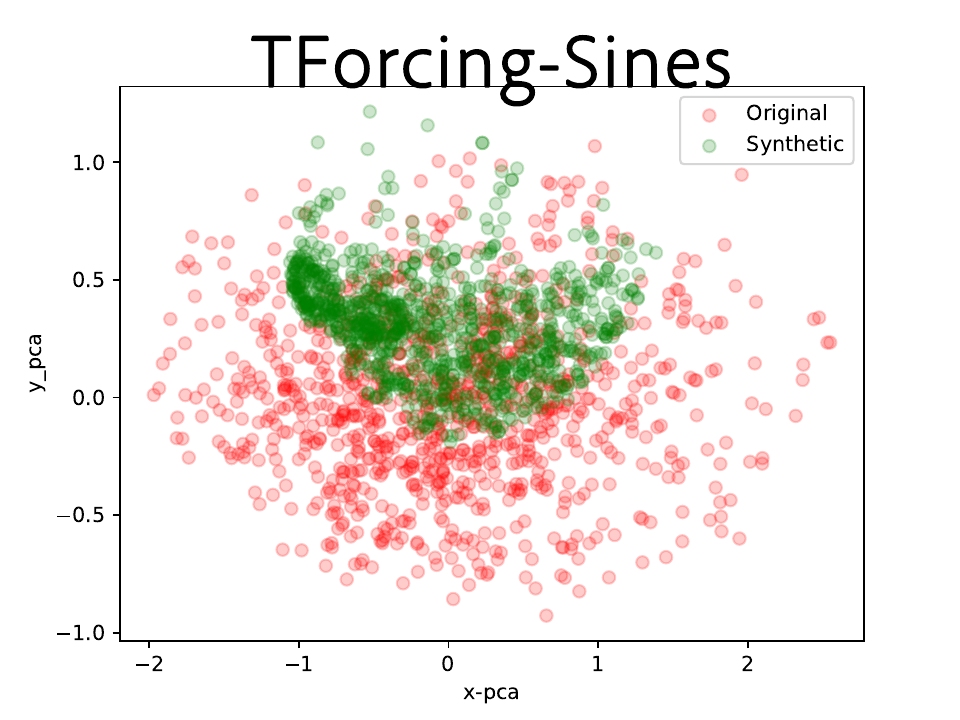}}\hfill
\subfloat{\includegraphics[width=0.122\textwidth]{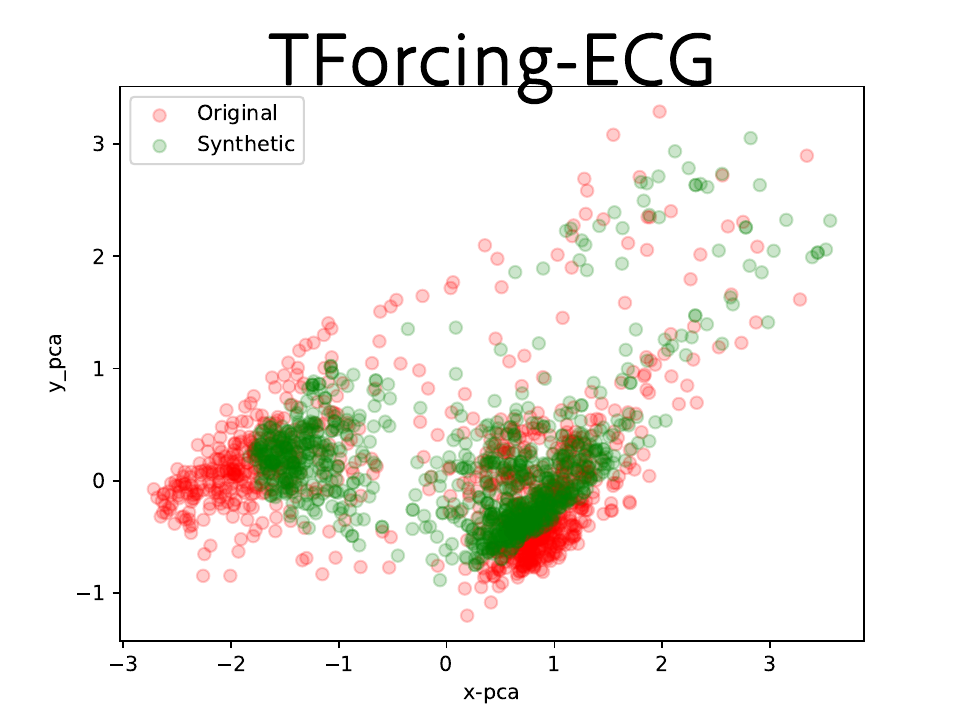}}\hfill
\subfloat{\includegraphics[width=0.122\textwidth]{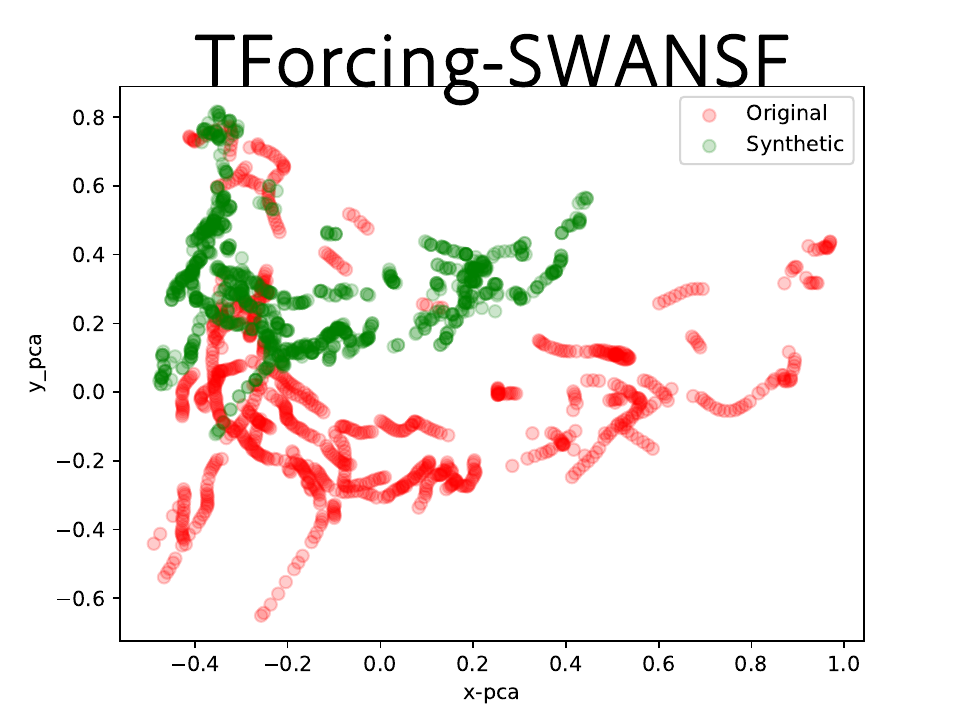}}\hfill

\vspace{-0.2cm}

\subfloat{\includegraphics[width=0.122\textwidth]{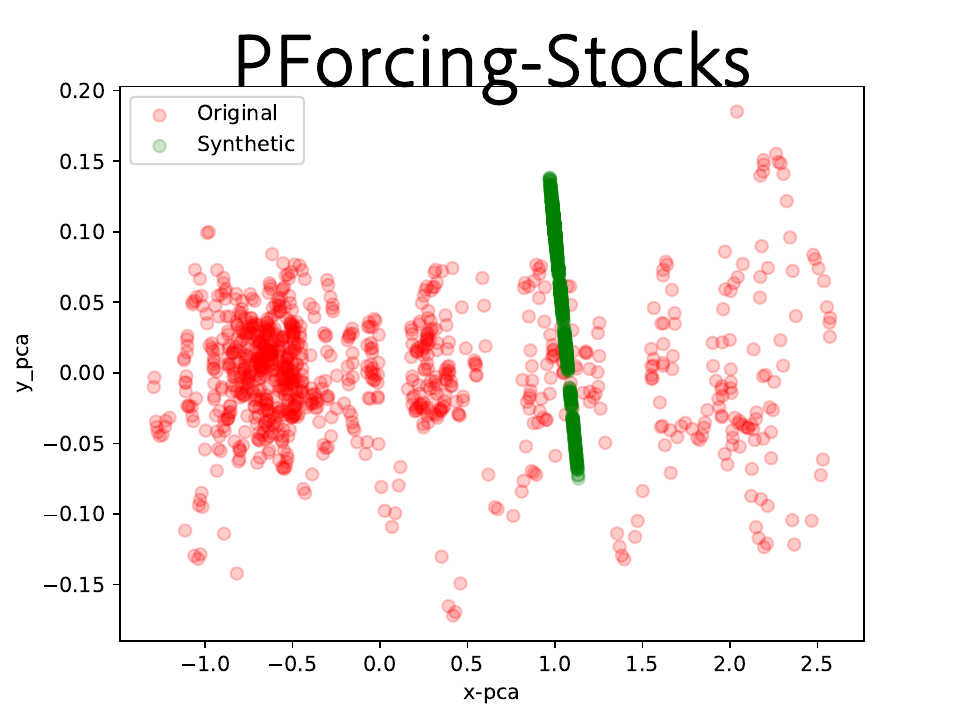}}\hfill
\subfloat{\includegraphics[width=0.122\textwidth]{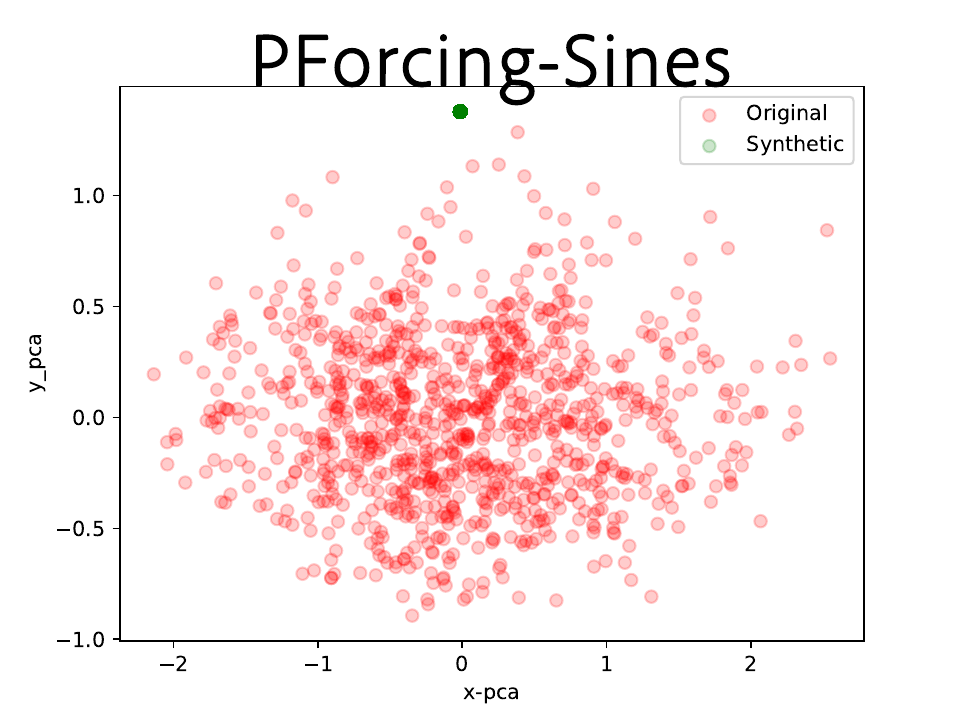}}\hfill
\subfloat{\includegraphics[width=0.122\textwidth]{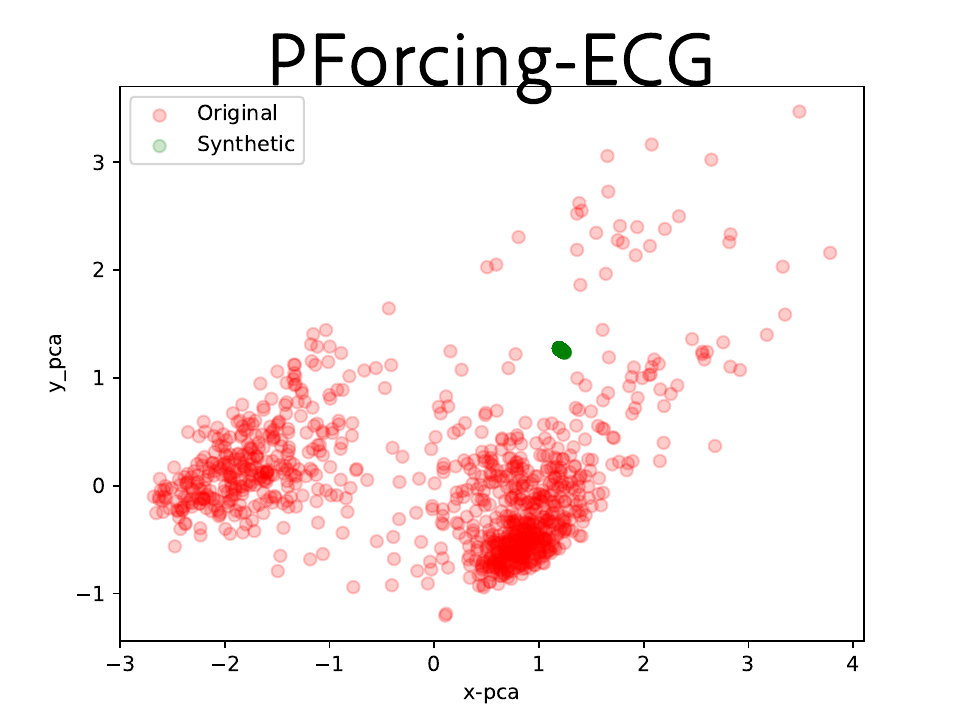}}\hfill
\subfloat{\includegraphics[width=0.122\textwidth]{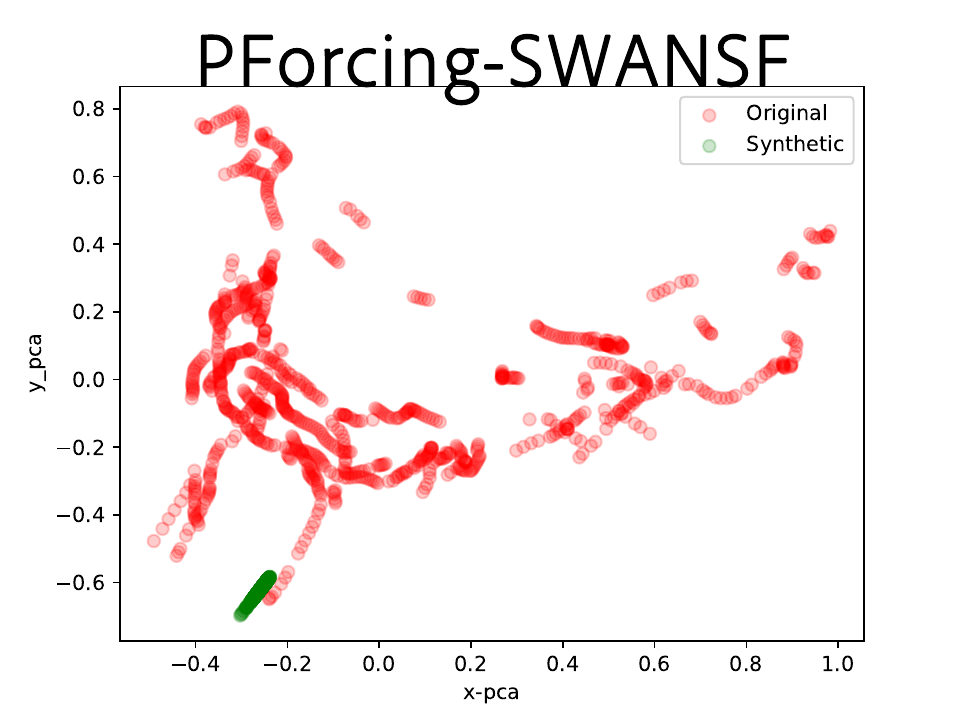}}\hfill

% Repeat for the second row and onwards
\caption{PCA visualizations illustrate how the distributions of original and synthetic data align. The top row shows SeriesGAN results, with TimeGAN, Standard GAN, Teacher Forcing, and Professor Forcing visualizations displayed sequentially underneath. From left to right, the plots correspond to the Stocks, Sines, ECG, and SWAN-SF datasets.}
\label{fig:pca}
\end{figure}

\begin{figure}
\centering
\subfloat{\includegraphics[width=0.122\textwidth]{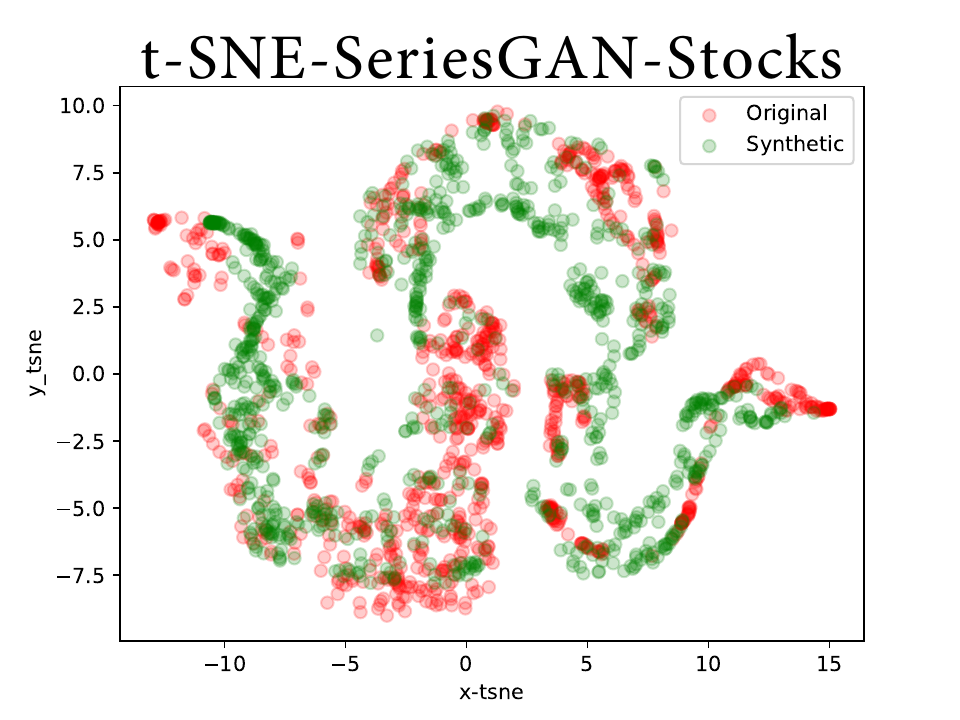}}\hfill
\subfloat{\includegraphics[width=0.122\textwidth]{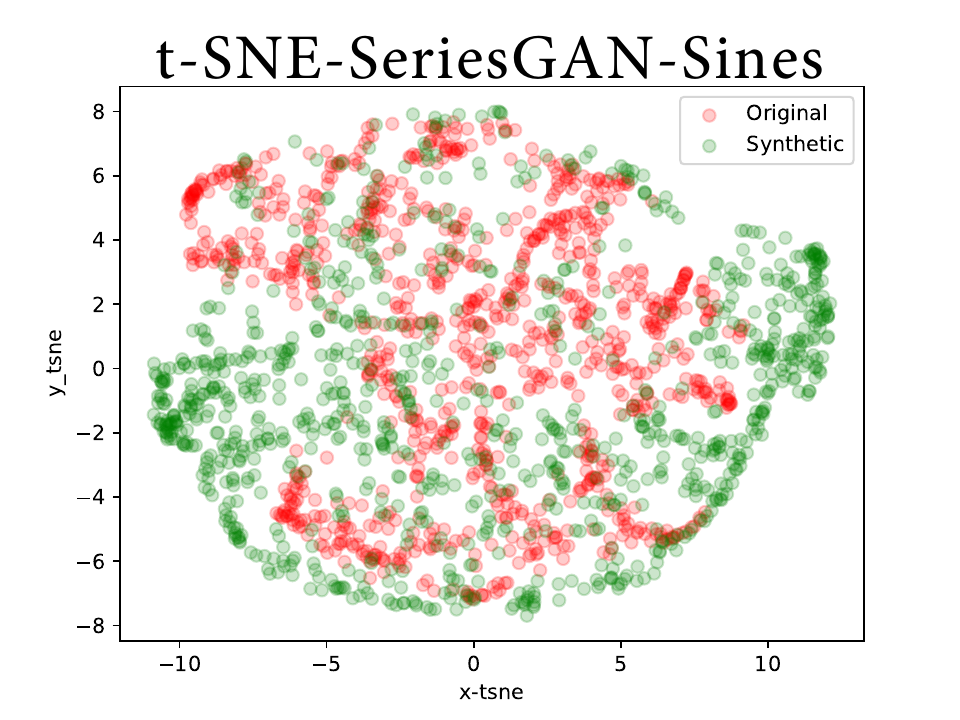}}\hfill
\subfloat{\includegraphics[width=0.122\textwidth]{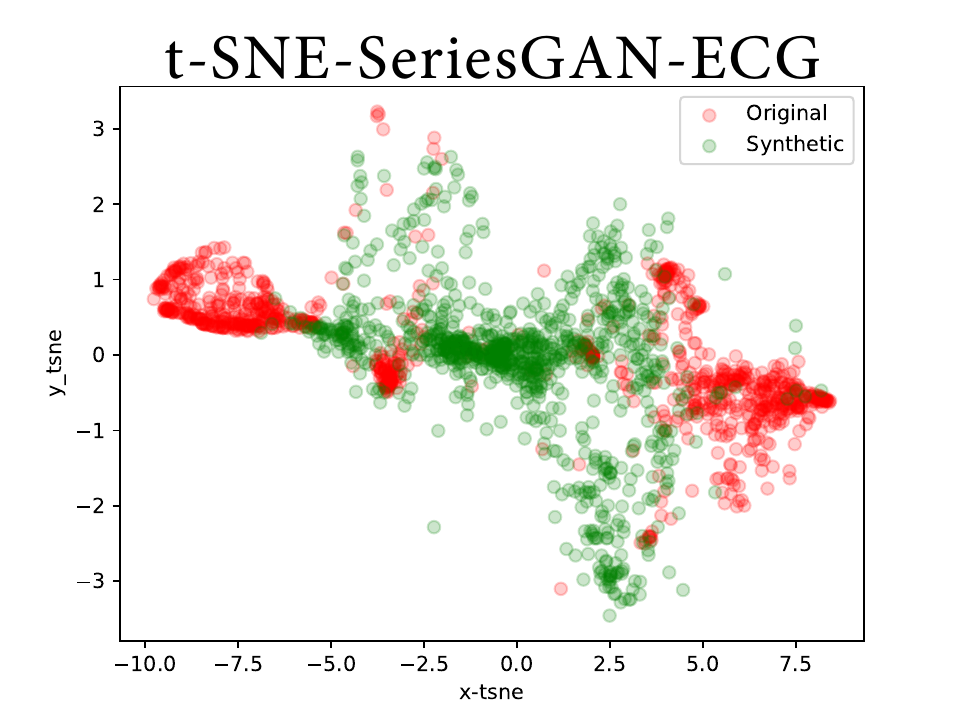}}\hfill
\subfloat{\includegraphics[width=0.122\textwidth]{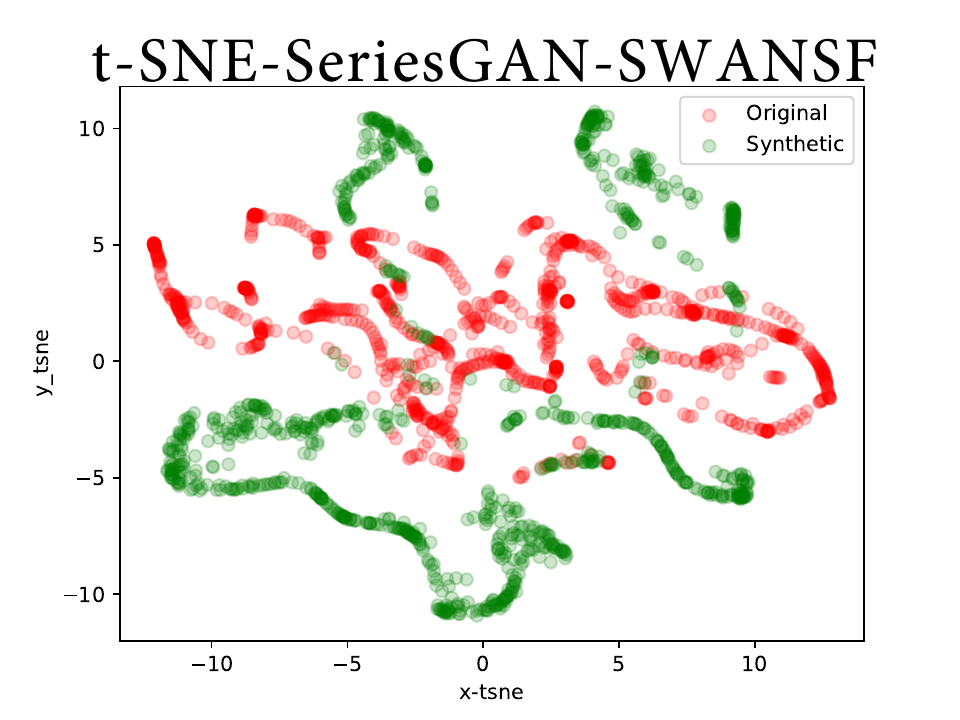}}\hfill

\vspace{-0.2cm}

\subfloat{\includegraphics[width=0.122\textwidth]{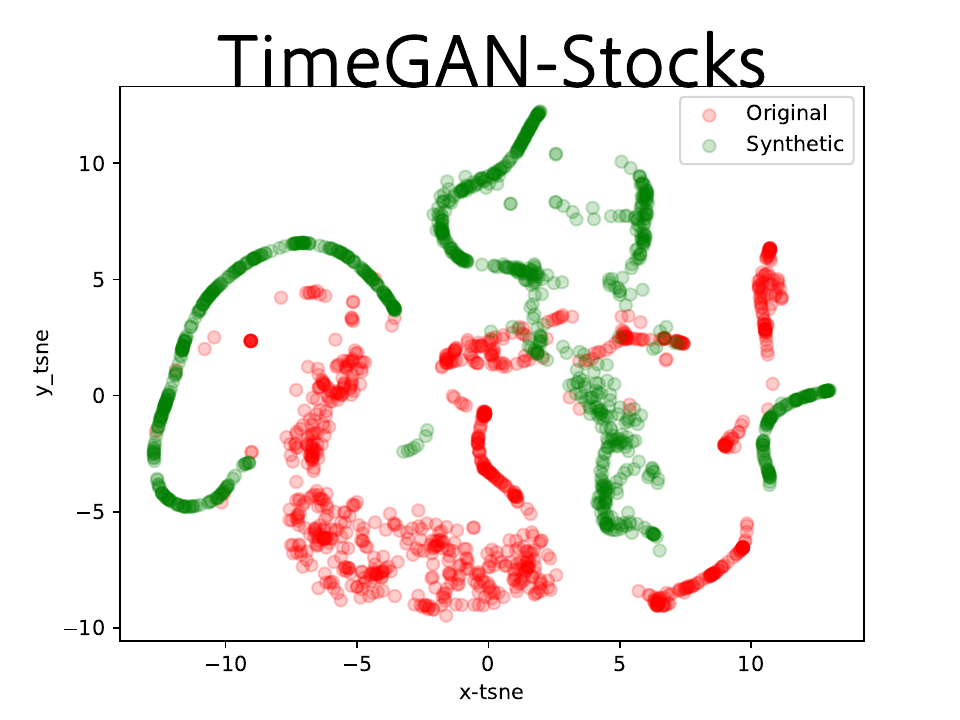}}\hfill
\subfloat{\includegraphics[width=0.122\textwidth]{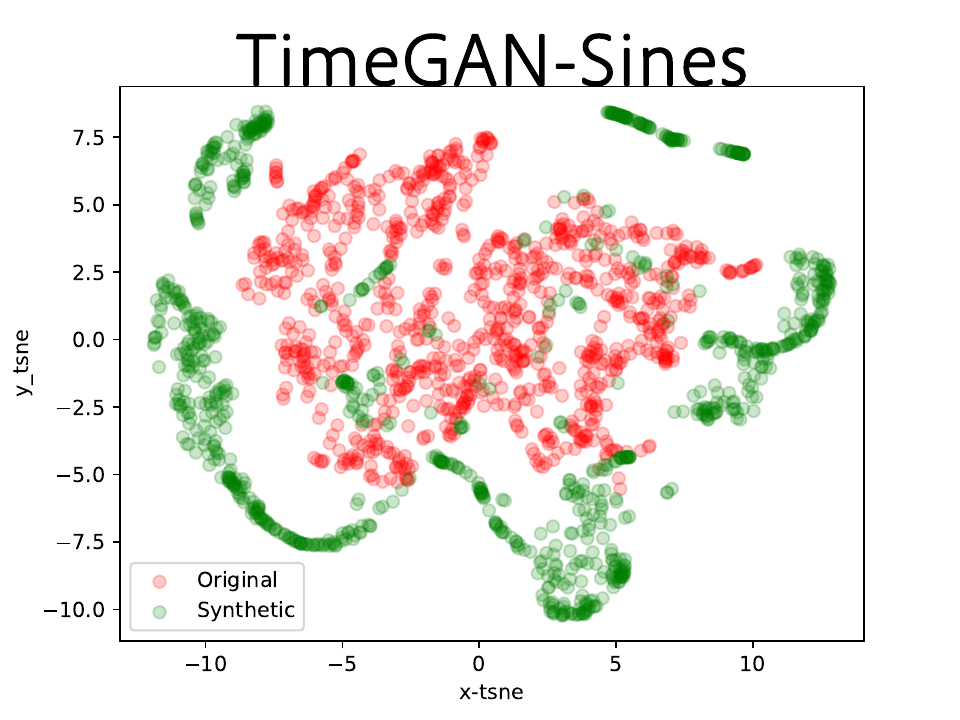}}\hfill
\subfloat{\includegraphics[width=0.122\textwidth]{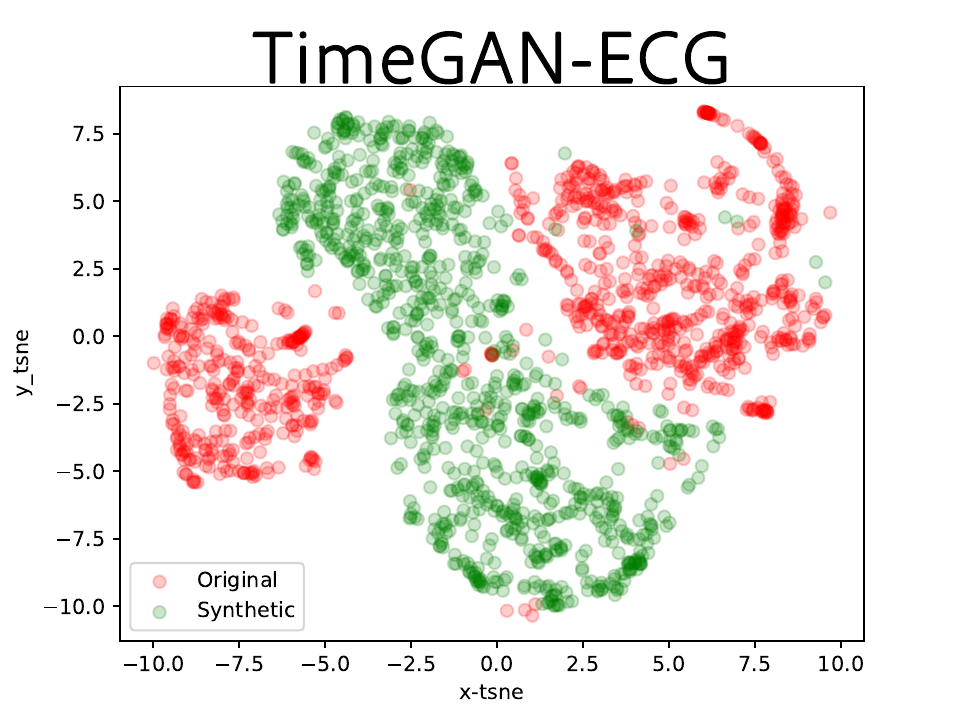}}\hfill
\subfloat{\includegraphics[width=0.122\textwidth]{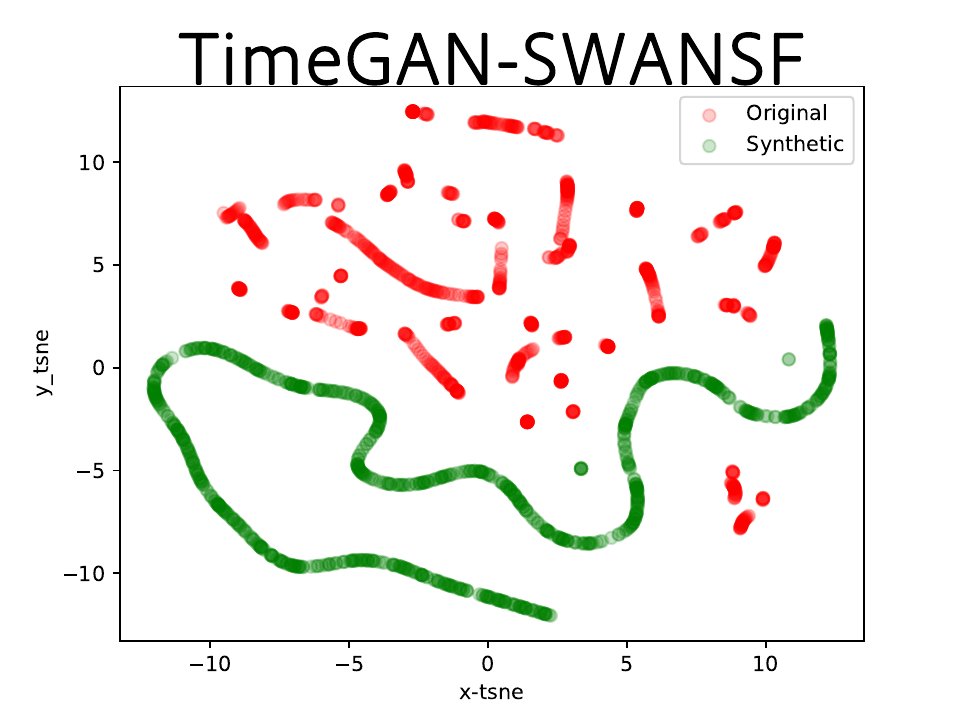}}\hfill

\vspace{-0.2cm}

\subfloat{\includegraphics[width=0.122\textwidth]{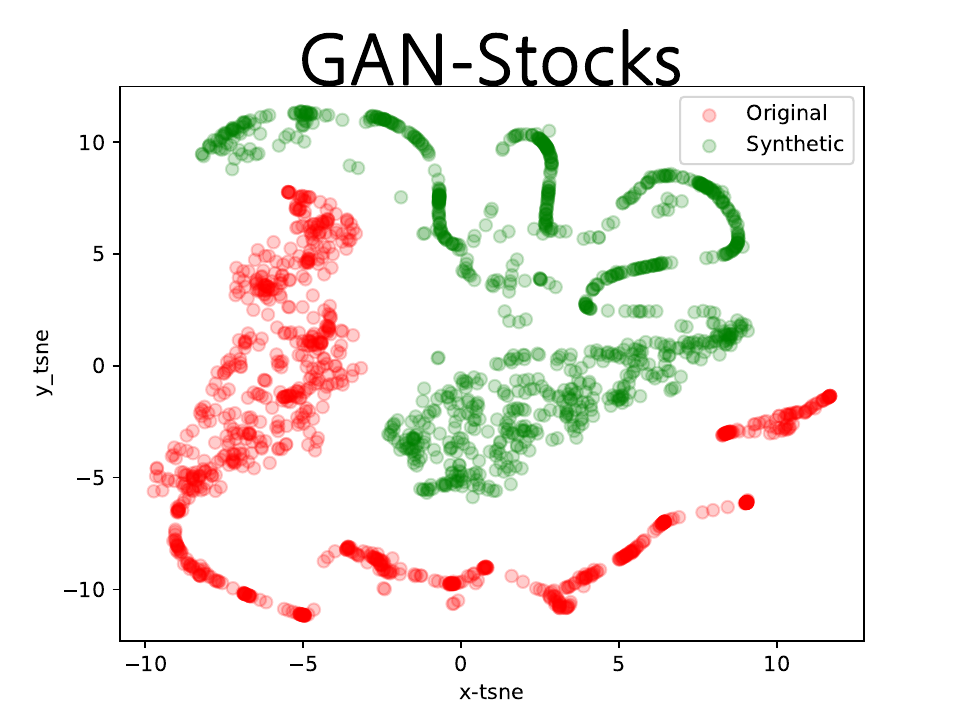}}\hfill
\subfloat{\includegraphics[width=0.122\textwidth]{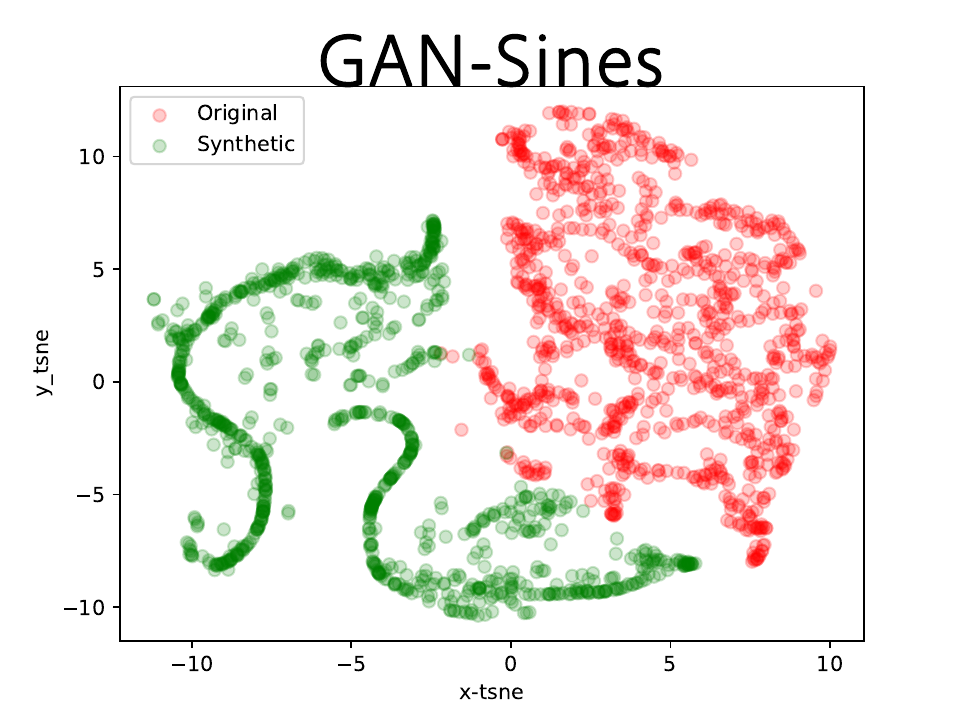}}\hfill
\subfloat{\includegraphics[width=0.122\textwidth]{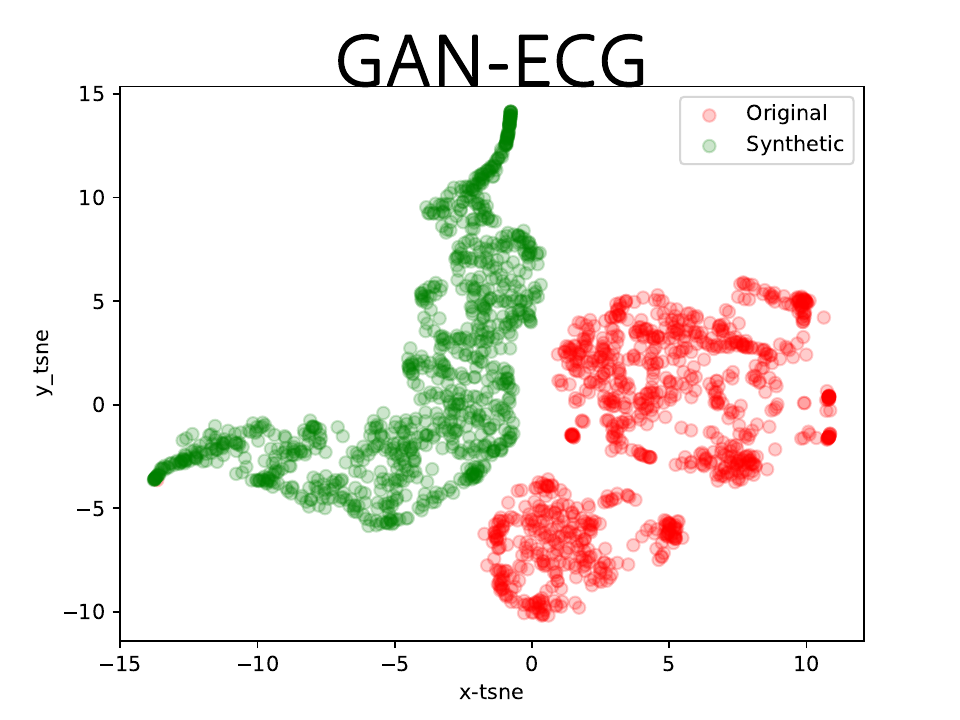}}\hfill
\subfloat{\includegraphics[width=0.122\textwidth]{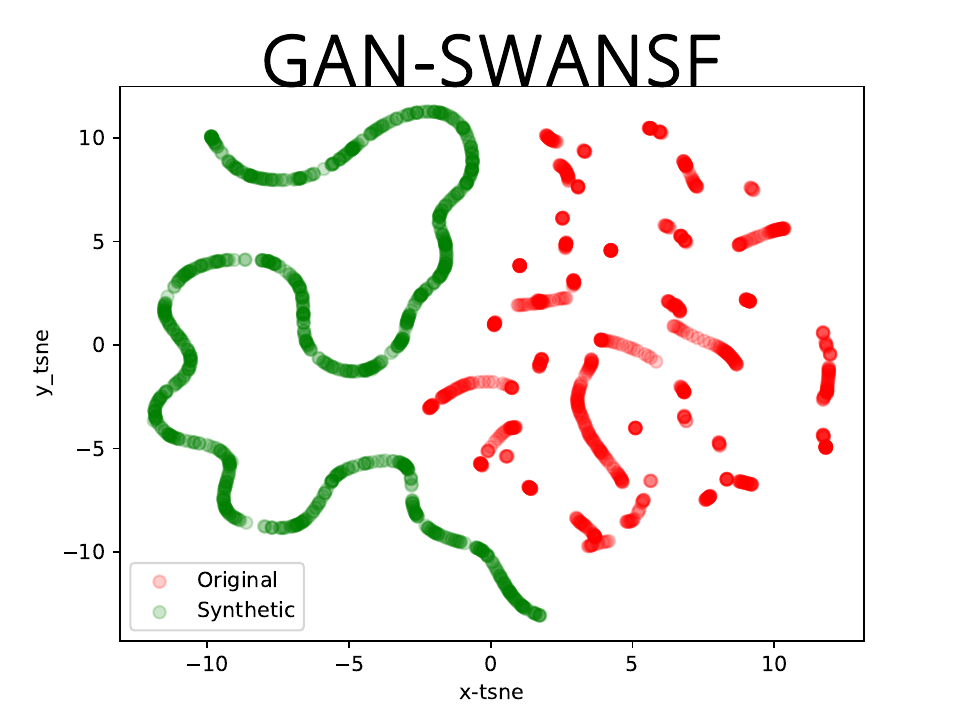}}\hfill

\vspace{-0.2cm}

\subfloat{\includegraphics[width=0.122\textwidth]{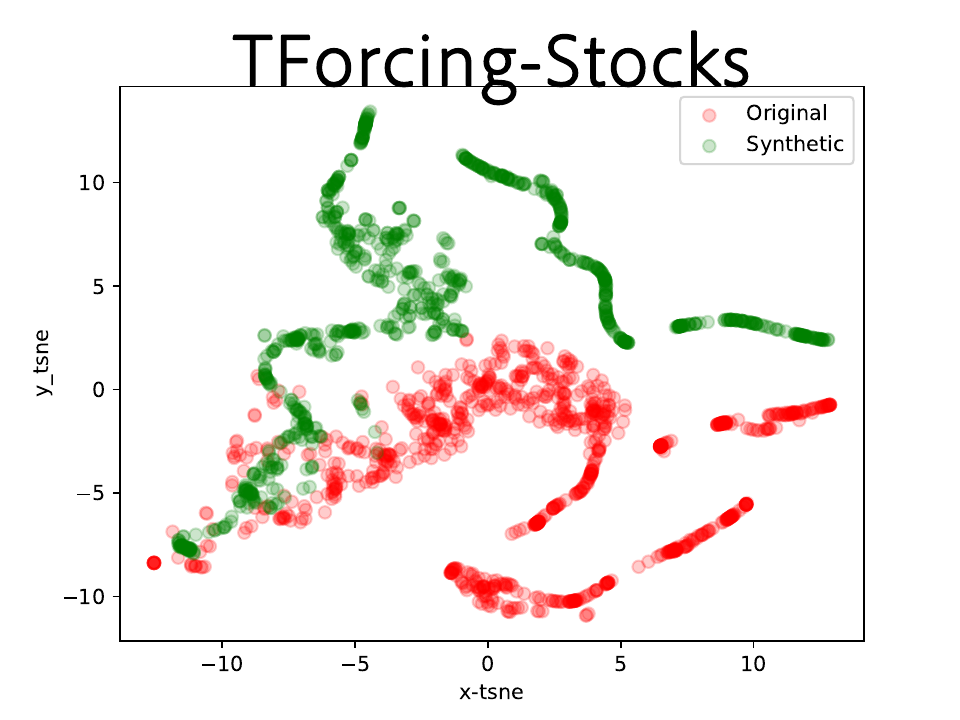}}\hfill
\subfloat{\includegraphics[width=0.122\textwidth]{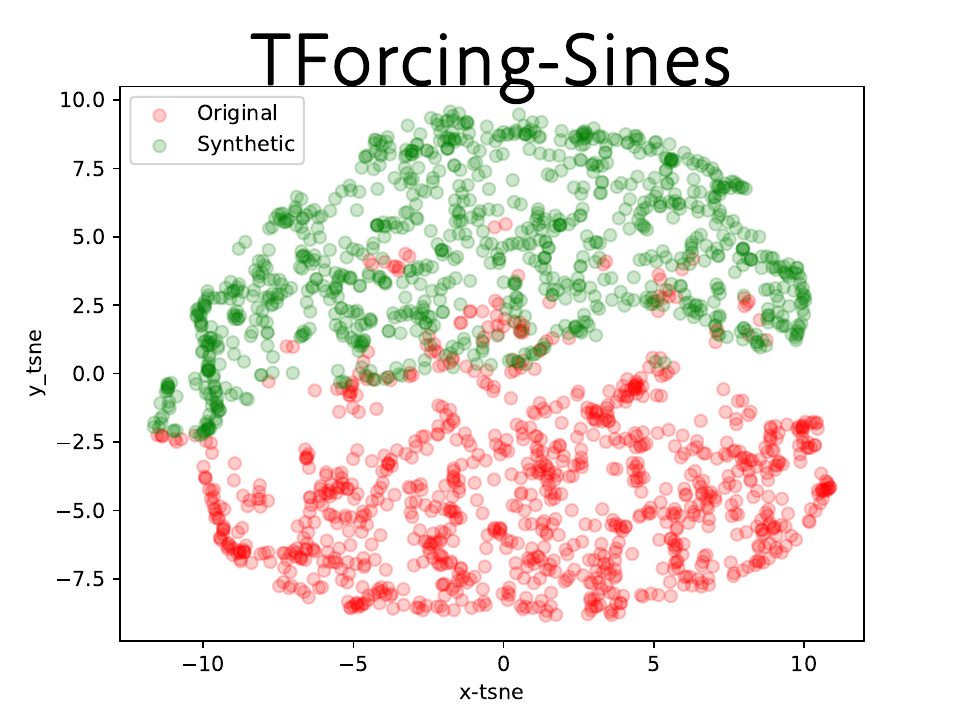}}\hfill
\subfloat{\includegraphics[width=0.122\textwidth]{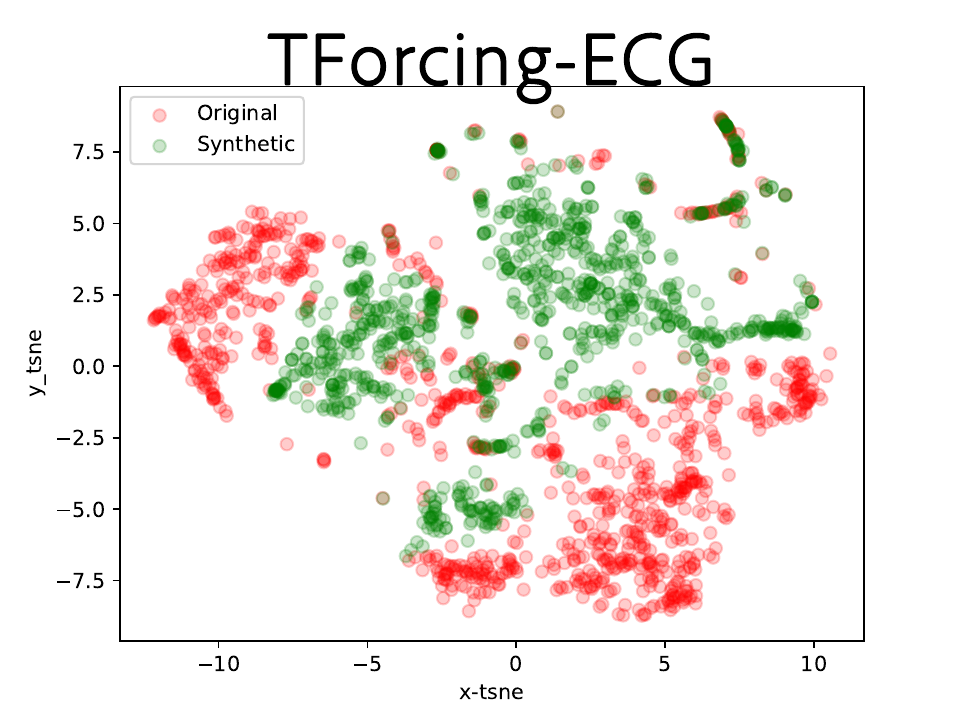}}\hfill
\subfloat{\includegraphics[width=0.122\textwidth]{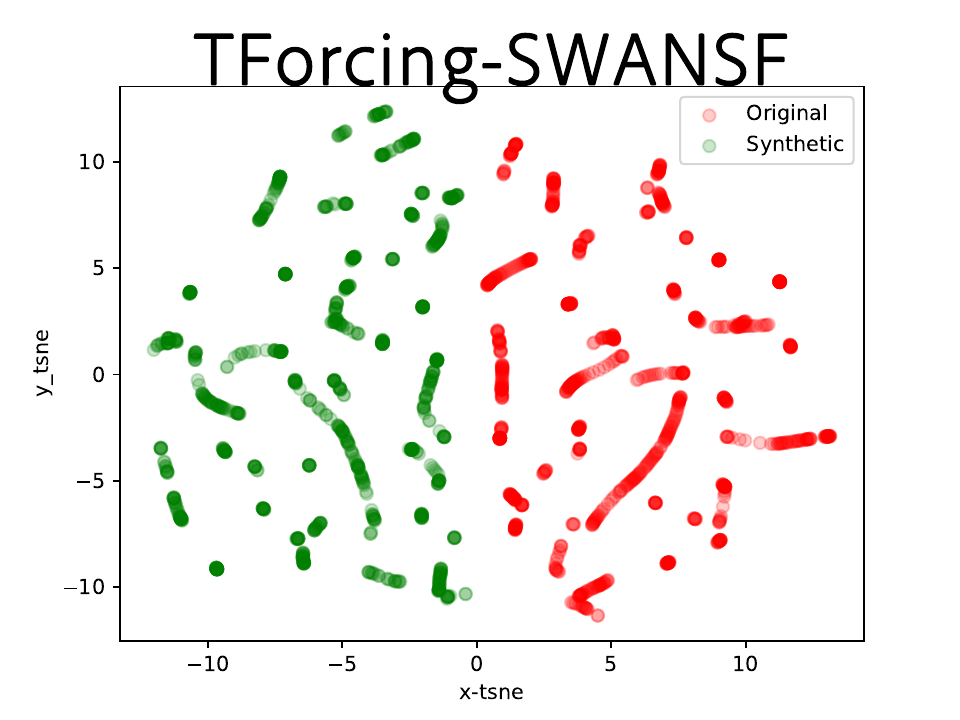}}\hfill

\vspace{-0.2cm}

\subfloat{\includegraphics[width=0.122\textwidth]{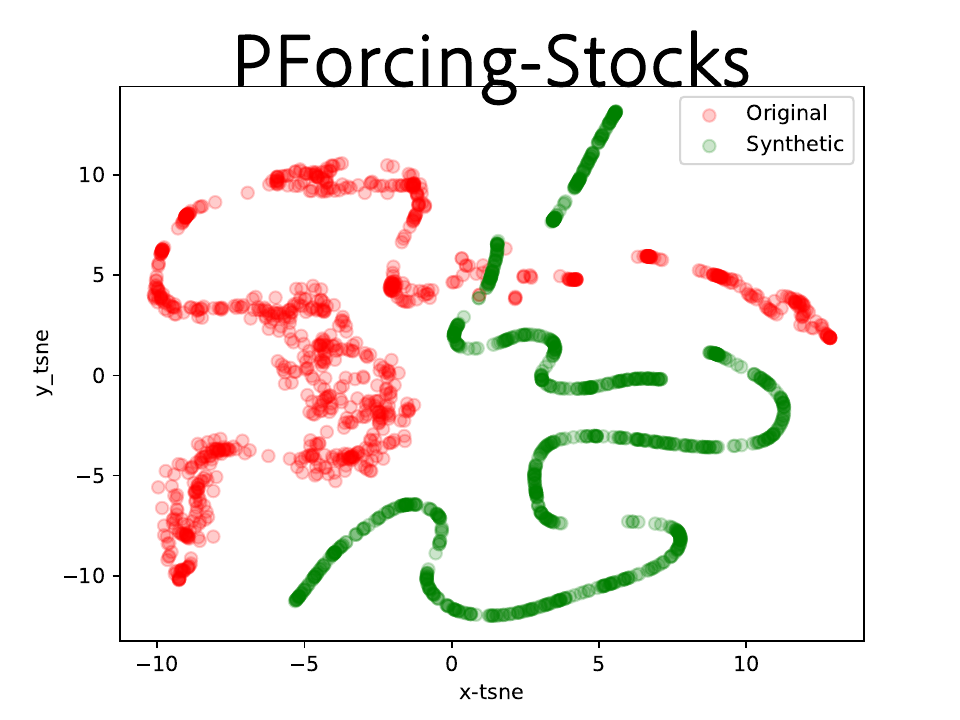}}\hfill
\subfloat{\includegraphics[width=0.122\textwidth]{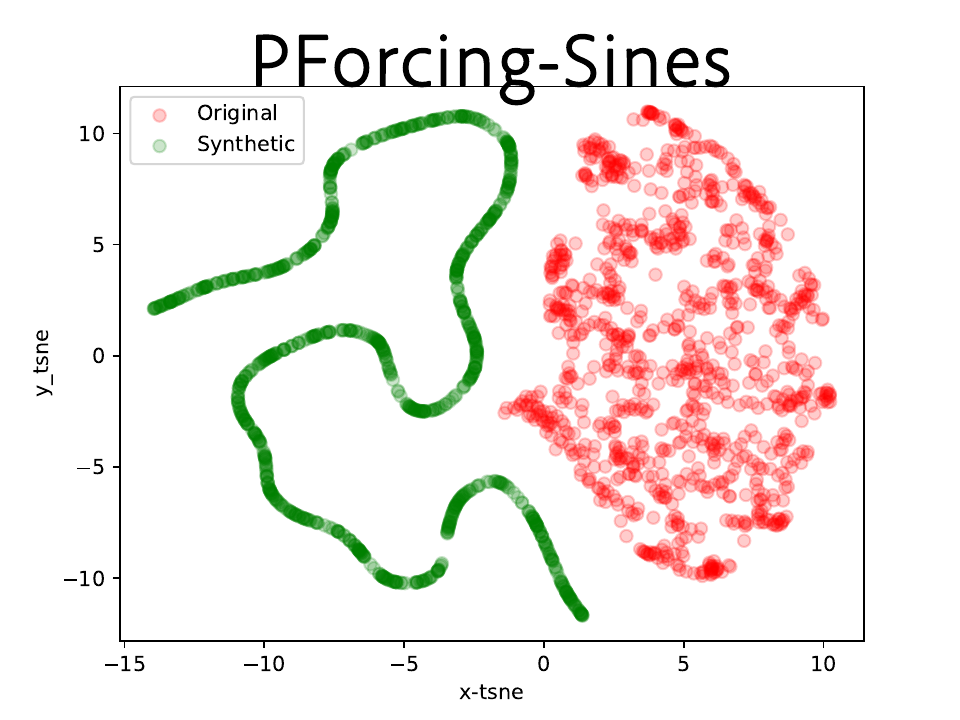}}\hfill
\subfloat{\includegraphics[width=0.122\textwidth]{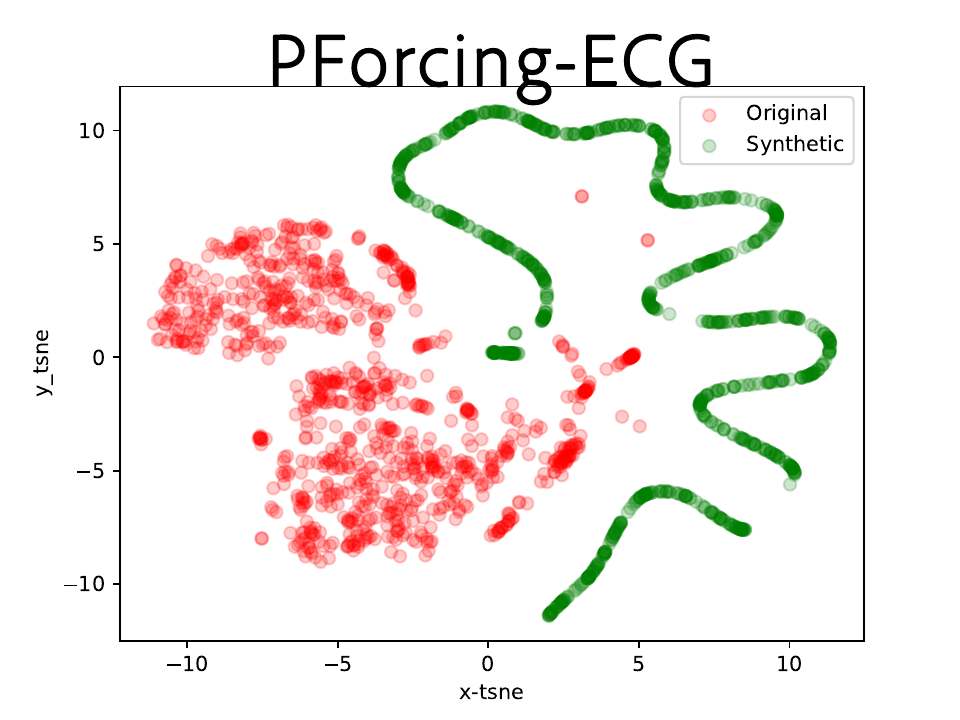}}\hfill
\subfloat{\includegraphics[width=0.122\textwidth]{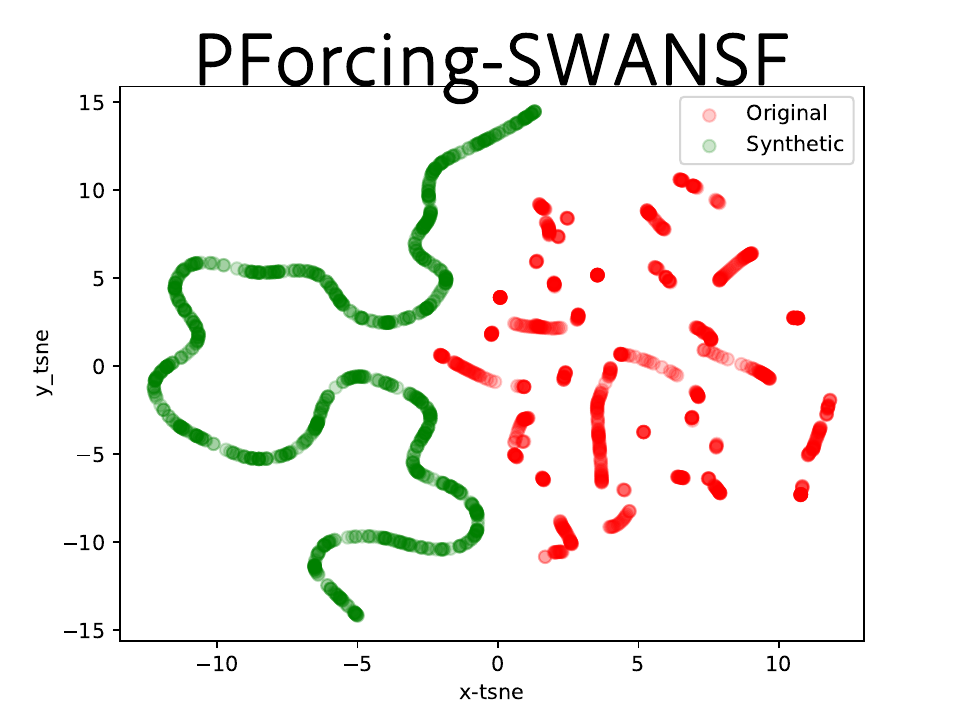}}\hfill

% Repeat for the second row and onwards
\caption{t-SNE visualizations show distribution alignment between original and synthetic data. The top row presents SeriesGAN results, followed by TimeGAN, Standard GAN, Teacher Forcing, and Professor Forcing from top to bottom. Left to right, plots represent Stocks, Sines, ECG, and SWAN-SF datasets.}

\label{fig:tsne}
\end{figure}

However, the analysis of t-SNE and PCA is also necessary to make a final decision, as a model may only learn part of a dataset’s probability distribution and generate high-quality data for just a portion of that distribution. As a result, high discriminative and predictive scores may be obtained, but the technique may not be as effective in learning the entire distribution. Based on Figs. \ref{fig:pca} and \ref{fig:tsne}, SeriesGAN is capable of learning the full data distribution and generating data across the entire distribution. Specifically, the PCA visualization results for the Stocks and Sines datasets are impressive compared to the baseline techniques, and the t-SNE visualization results for Stocks and Sines are also promising.

Therefore, based on the three evaluation metrics and the comparison of results, SeriesGAN achieved the desired objectives discussed in Section \ref{sec:problem}. It also outperformed the best state-of-the-art techniques by a significant margin in both quantitative and qualitative terms. However, as shown in Figs. \ref{fig:pca} and \ref{fig:tsne}, the primary weakness of existing time series generation techniques, including SeriesGAN, is their suboptimal performance on long series, in this case the ECG and SWAN-SF datasets. We define “long series” as those with a number of timestamps equal to or exceeding 60. These GAN-based techniques encounter difficulties in learning temporal dynamics over extended durations.

\begin{table*}
\centering
\caption{The contribution of each innovation in SeriesGAN is evaluated in terms of both discriminative and predictive scores}
\label{tbl:contribution}
\begin{tabular}{cccc}
\hline
\multicolumn{4}{c}{\textit{Discriminative Score}} \\ \hline
\textbf{} & \textbf{Stocks} & \textbf{Sines} & \textbf{ECG}\\ \hline

\textbf{SeriesGAN} & \textbf{0.1873 ± 0.0823} & \textbf{0.2083 ± 0.0869} & \textbf{0.1691 ± 0.0234} \\ 

w/o Novel Supervised Loss & 0.2334 ± 0.0441 & 0.2732 ± 0.0949 & 0.2427 ± 0.0581 \\

w/o Dual Discriminators & 0.2119 ± 0.0043 & 0.2340 ± 0.0886 & 0.2284 ± 0.0201 \\ 

w/o Autoencoder-based Time Series Loss & 0.2265 ± 0.0383 & 0.2497 ± 0.0991 & 0.1921 ± 0.0487 \\

w/o Early Stopping & 0.2782 ± 0.0443 & 0.2563 ± 0.1042 & 0.2677 ± 0.1003 \\ \hline

\multicolumn{4}{c}{\textit{Predictive Score}} \\ \hline

\textbf{} & \textbf{Stocks} & \textbf{Sines} & \textbf{ECG} \\ \hline

\textbf{SeriesGAN} & \textbf{0.041 ± 0.0002} & \textbf{0.2232 ± 0.0018} & \textbf{0.1268 ± 0.0007} \\ 

w/o Novel Supervised Loss & 0.0844 ± 0.0069 & 0.2792 ± 0.027 & 0.1303 ± 0.0022 \\ 

w/o Dual Discriminators & 0.0419 ± 0.0088 & 0.2306 ± 0.0199 & 0.1533 ± 0.008 \\ 

w/o Autoencoder-based Time Series Loss & 0.0594 ± 0.0081 & 0.261 ± 0.0072 & 0.1391 ± 0.0037 \\ 

w/o Early Stopping & 0.0955 ± 0.0203 & 0.259 ± 0.0122 & 0.1897 ± 0.0361 \\ \hline

\end{tabular}
\end{table*}

\subsection{Contribution of Each Novelty}

In this section, we analyze the contribution and impact of each novel component integrated into the SeriesGAN framework. First, we compare SeriesGAN with a variant that excludes the novel supervised loss used for autoregressive learning, allowing us to assess the significance of this feature. Next, we investigate the effect of removing the dual discriminator mechanism, retaining only the latent discriminator to evaluate its influence on network performance. Additionally, we eliminate the autoencoder-based loss function to observe its specific impact on the network’s behavior. Finally, we disable the early stopping algorithm to examine the resulting changes in overall performance. Each modification helps isolate and understand the role of these components in enhancing the SeriesGAN framework.

As shown in Table \ref{tbl:contribution}, the novel supervised loss and early stopping significantly enhance both discriminative and predictive scores. Specifically, incorporating the novel supervised loss led to a 33.1\% reduction in discriminative score and a 44.57\% reduction in predictive score. Similarly, implementing early stopping yielded substantial improvements, reducing the discriminative and predictive scores by 43.29\% and 66.19\%, respectively. In contrast, dual discriminator training had a relatively modest impact on metric improvement. Meanwhile, the autoencoder-based time series loss demonstrated a promising effect on enhancing evaluation metrics, contributing to the quality of the generated data. Ultimately, the combination of these innovations creates a robust and stable network capable of generating high-quality time series data while effectively capturing the underlying distribution of the real data.

\subsection{Examples of Generated Data}

\begin{figure}
\centering
\includegraphics[width=0.45\textwidth]{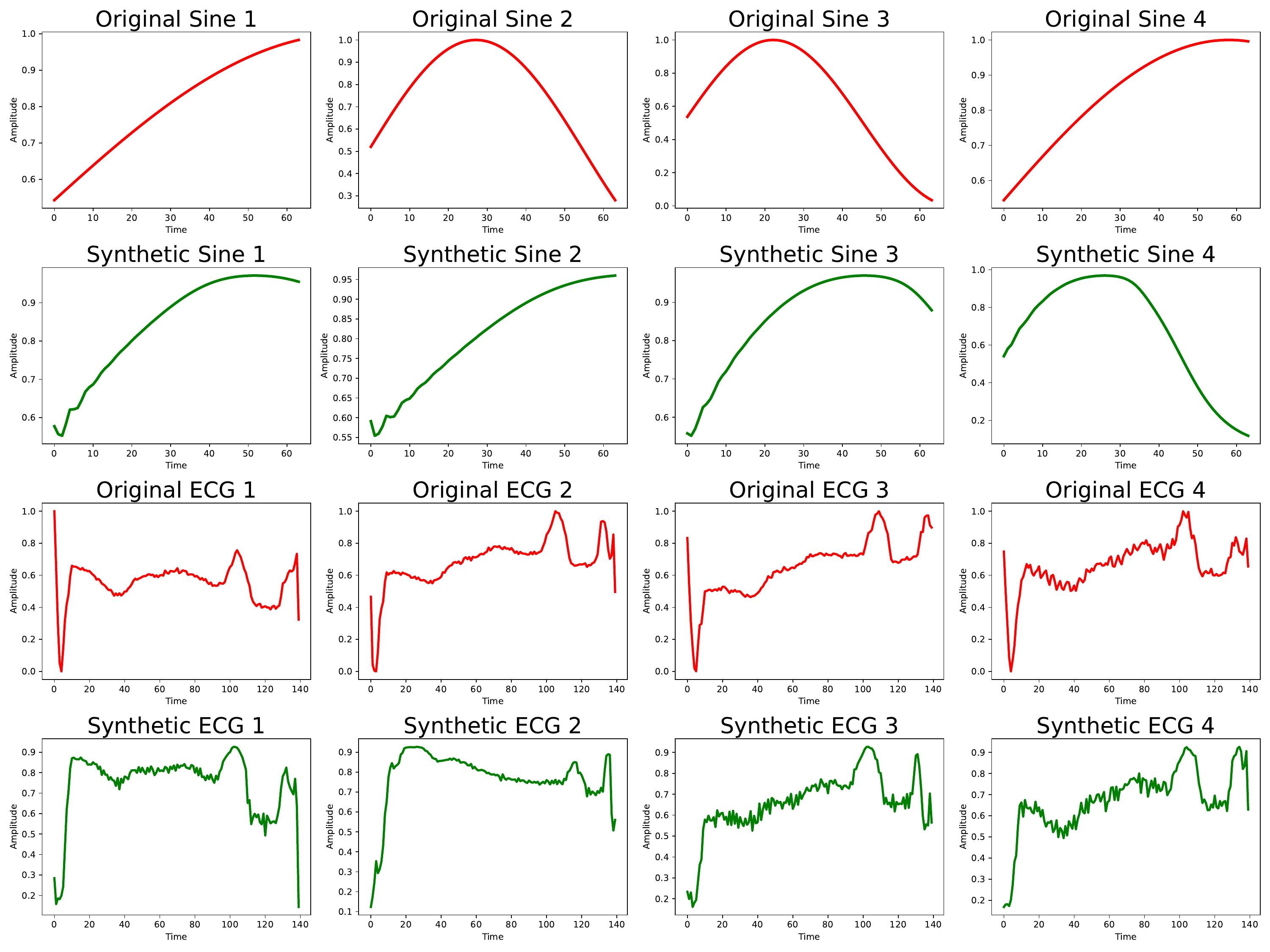}
\caption{This illustration compares the original dataset samples (red) with their synthetic counterparts generated by the SeriesGAN algorithm (green) for both Sines and ECG samples.}
\label{fig:example}
\end{figure}

In this section, we present four randomly selected real and synthetic data samples generated by SeriesGAN from the Sines and ECG datasets, both of which exhibit consistent temporal dynamics throughout their samples. This makes it meaningful and easy to compare them without requiring specific knowledge of the dataset. These examples are shown in Fig. \ref{fig:example}. The results demonstrate SeriesGAN’s effectiveness in learning the temporal dynamics of time series data. For the Sines dataset, which is a multivariate time series, only one feature is displayed. The results of the generated Sines samples display smooth data without any noise, demonstrating the effectiveness of combining a GAN network with autoregressive learning. This integration enables the framework to effectively capture the temporal dynamics of the data. Furthermore, this success is attributed to the novel loss function introduced for teacher forcing training, which enhances the model’s ability to learn temporal dependencies. By focusing on the second subsequent time step instead of just the next, the model gains a more comprehensive understanding of the temporal structure in the time series data.

\section{Conclusion}
\label{sec:conclusion}

In this study, we introduce SeriesGAN, an innovative model designed for generating high-quality time series data. SeriesGAN surpasses TimeGAN and other advanced methods by incorporating novel autoregressive training and a dual-discriminator approach. These enhancements significantly improve the model’s ability to capture temporal dynamics, reduce information loss in the embedding space, and strengthen the overall effectiveness of adversarial training. This improvement significantly boosts the performance of the autoencoder and generator networks. Additionally, SeriesGAN introduces a novel autoencoder-based loss function, employs Least Squares GANs, and includes an early stopping mechanism. This framework consistently surpasses current leading methods in generating realistic time series data. Future work will explore integrating these concepts into adversarial autoencoders to further develop a framework for producing high-quality time series data. Another valuable research direction is the incorporation of window-based (or kernel-based) learning into these GAN models. This approach could pave the way for new techniques that are better suited for generating long sequences.

\section{Acknowledgment}

This work has been supported by the Division of Atmospheric and Geospace Sciences within the Directorate for Geosciences through NSF awards \#2301397, \#2204363, and \#2240022, as well as by the Office of Advanced Cyberinfrastructure within the Directorate for Computer and Information Science and Engineering under NSF award \#2305781.

\balance

\end{document}